\documentclass{article}

\usepackage[final]{neurips_2024}

\usepackage[utf8]{inputenc} % allow utf-8 input
\usepackage[T1]{fontenc}    % use 8-bit T1 fonts
\usepackage{hyperref}       % hyperlinks
\usepackage{url}            % simple URL typesetting
\usepackage{booktabs}       % professional-quality tables
\usepackage{amsfonts}       % blackboard math symbols
\usepackage{nicefrac}       % compact symbols for 1/2, etc.
\usepackage{microtype}      % microtypography
\usepackage{xcolor}         % colors
\usepackage{times}
\usepackage{latexsym}
\usepackage[T1]{fontenc}
\usepackage[utf8]{inputenc}
\usepackage{microtype}
\usepackage{inconsolata}
\usepackage{lipsum}
\usepackage{stfloats}
\usepackage{graphicx}
\usepackage{paralist}
\usepackage{amsmath}
\usepackage{multirow}
\usepackage{caption}
\usepackage{subcaption}
\usepackage{xspace}
\usepackage{makecell}
\usepackage{algorithm}  
\usepackage{algorithmic}
\usepackage{listings}
\usepackage{url}
\usepackage{booktabs}
\usepackage{colortbl}
\usepackage{bbm}
\usepackage{amssymb}
\usepackage{float}
\usepackage[capitalise]{cleveref}
\usepackage{comment}
\usepackage{fontawesome5}
\usepackage{academicons}

\newcommand{\norm}{\(L_2\) norm\xspace}
\newcommand{\kvcache}{KV Cache\xspace}
\newcommand{\corrscore}{\textsc{ALr}\xspace}

\definecolor{darkblue}{rgb}{0, 0, 0.5}
\hypersetup{colorlinks=true, citecolor=darkblue, linkcolor=darkblue, urlcolor=darkblue}

\author{
\textbf{Alessio Devoto}\textsuperscript{Q}\footnotemark[1]\qquad
\textbf{Yu Zhao}\textsuperscript{K}\footnotemark[1]\qquad
\textbf{Simone Scardapane}\textsuperscript{Q}\qquad
\textbf{Pasquale Minervini}\textsuperscript{K,V}\\
\textsuperscript{Q}Sapienza University of Rome\qquad \textsuperscript{K}The University of Edinburgh \qquad \textsuperscript{V}Miniml.AI \\
\texttt{\{alessio.devoto, simone.scardapane\}@uniroma1.it} \\
\texttt{\{yu.zhao, p.minervini\}@ed.ac.uk}  \\
\footnotesize
{\faGithub}\hspace{0.5em}\texttt{\url{https://github.com/alessiodevoto/l2compress}}
}

\title{A Simple and Effective \(L_2\) Norm-Based Strategy \\ for \kvcache Compression}

\begin{document}
\maketitle
\def\thefootnote{*}\footnotetext[1]{Equal contribution.}

\begin{abstract}
The deployment of large language models (LLMs) is often hindered by the extensive memory requirements of the Key-Value (KV) cache, especially as context lengths increase.
Existing approaches to reduce the \kvcache size involve either fine-tuning the model to learn a compression strategy or leveraging attention scores to reduce the sequence length.
We analyse the attention distributions in decoder-only Transformers-based models and observe that attention allocation patterns stay consistent across most layers. 
Surprisingly, we find a clear correlation between the \norm and the attention scores over cached KV pairs, where a low \norm of a key embedding usually leads to a high attention score during decoding.
This finding indicates that \emph{the influence of a KV pair is potentially determined by the key embedding itself before being queried}.
Based on this observation, we compress the \kvcache based on the \norm of key embeddings.
Our experimental results show that this simple strategy can reduce the \kvcache size by 50\% on language modelling and needle-in-a-haystack tasks and 90\% on passkey retrieval tasks without losing accuracy.
Moreover, without relying on the attention scores, this approach remains compatible with FlashAttention, enabling broader applicability.
\end{abstract}
%

% \begin{figure}[t]
%     \centering
%     \includegraphics[width=\columnwidth]{figures/norm_attn_diff/layers_heads_norm_attn_diff-crop.pdf}
%     \caption{\corrscore , as defined in \Cref{eq:norm-attn-diff}, for each head and layer in Llama2-7b. A lower value means a higher correlation between \norm and attention score.}
%     \label{fig:norm-attn-diff}
% \end{figure}

%
\section{Introduction}
Handling long contexts is desirable for large language models (LLMs), as it allows them to perform tasks that require understanding long-term dependencies~\cite{liu-etal-2024-lost,yaofu-data-eng,longlora,structured-packing,zhao2024analysing,longllama}.
A key component for modelling long context is the \kvcache, which stores the keys and values of past tokens in memory to avoid recomputing them during generation.
However, processing long-context inputs often results in a high decoding latency since it requires repeatedly reading a potentially large \kvcache from high-bandwidth memory (HBM) to the streaming multiprocessor (SM) during decoding~\citep{yaofu-long-context-challenge}.
Consequently, the practical deployment of LLMs is frequently hindered by hardware limitations.
To address the issue of \kvcache growth, various \kvcache compression methods have been proposed.
These methods can be broadly categorised into trainable approaches, which involve modifications to the model architecture~\cite{ainslie2023gqa}, or fine-tuning regime to inherently manage \kvcache size~\cite{nawrot2024dynamic}, and non-trainable approaches, which apply post-hoc compression techniques to reduce the cache footprint without altering the underlying model \cite{li2024snapkv, h2o, ge2023model}.
While these methods have shown promise, they often involve complex algorithms or significant computational overhead, limiting their practicality; for example, post-hoc compression algorithms usually evict KV pairs based on attention scores, which is not compatible with FlashAttention~\citep{dao2022flashattention} and thus prevents their applications in modern LLMs inference systems.
We show that, surprisingly, the \norm of cached keys has a high correlation with attention scores.
More specifically, we observe that a low \norm of a key embedding usually leads to a high attention score during decoding.
Based on this observation, we propose a simple and highly effective strategy for \kvcache compression: \emph{keeping in memory only the keys with lowest \norm, and the corresponding values}.
Unlike many existing methods, our heuristic can be applied off-the-shelf to any transformer-based decoder-only LLM without the need for additional training or significant modifications.
More importantly, our method estimates the influence of cached key-value pairs without the need to compute the attention scores. Therefore, unlike other compression methods \citep{deepspeed-fastgen, li2024snapkv}, it can be easily integrated with the popular FlashAttention \citep{dao2022flashattention}.
%thus potentially saving compute resources.
%

%
Our experimental results demonstrate that this heuristic allows maintaining model performance in language modelling tasks and in tasks that require the model to store and retrieve the most critical information, such as passkey retrieval~\citep{passkey-retrieval} and needle-in-a-haystack tasks~\citep{needleinhaystack}. 

\begin{figure*}[t]
  \centering
  \begin{subfigure}{\textwidth}
    \centering
    \includegraphics[width=\textwidth]{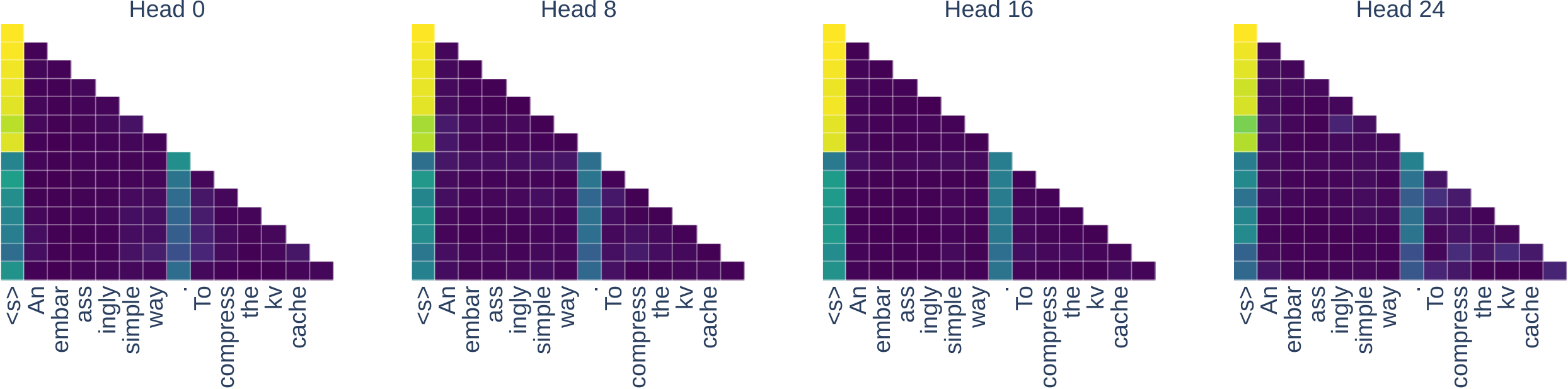}
    \label{fig:attention_scores}
  \end{subfigure}
  
  \begin{subfigure}{\textwidth}
    \centering
    \includegraphics[width=\textwidth]{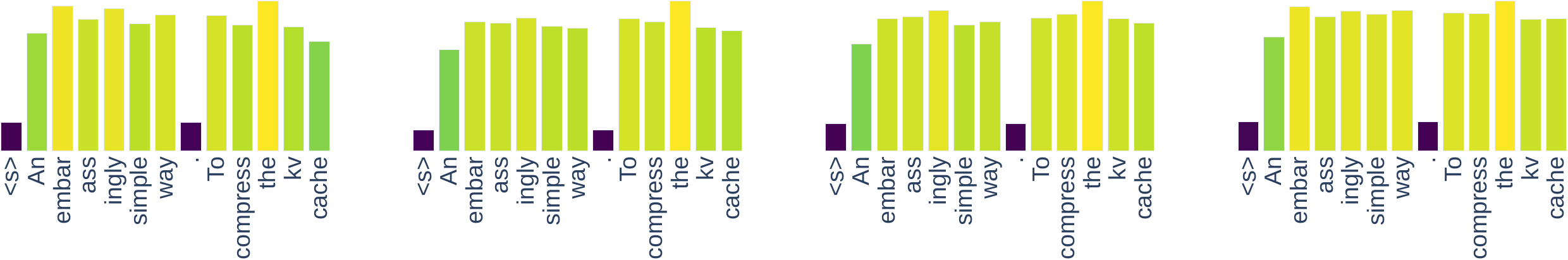}
    \label{fig:norms}
  \end{subfigure}
  \caption{Five heads at layer 9 of Llama2-7b. Attention score (top) and \(L_2\) norm (bottom)  are highly correlated.
  We observe similar patterns across most layers and for a wide range of inputs. More examples provided in \Cref{sec:more_visualizations}
  }
  \label{fig:attn_norms_kurt}
\end{figure*}

\section{Background on LLM Inference}
In transformer-based LLMs, the input sequence is represented as a tensor $\mathbf{X} \in \mathbb{R}^{n \times d}$, where $n$ is the sequence length and $d$ is the token embedding dimension. Each $x_i$ corresponds to an embedding of a token in the sequence. The tensor $\mathbf{X}$ is processed by a series of transformer blocks, each composed of a multi-head self-attention and a feed-forward layer.

Given an input $\mathbf{X} \in \mathbb{R}^{n \times d}$, the multi-head attention mechanism performs multiple attention operations in parallel, allowing the model to attend to information from different representation subspaces. It does so by first computing three projections: the query, key, and value matrices, denoted as $\mathbf{Q}$, $\mathbf{K}$, and $\mathbf{V}$, respectively. These are obtained by linear transformations of the input $\mathbf{X}$:
\begin{equation}
\mathbf{Q} = \mathbf{X} \mathbf{W}_Q, \quad \mathbf{K} = \mathbf{X} \mathbf{W}_K, \quad \mathbf{V} = \mathbf{X} \mathbf{W}_V,
\end{equation}
where $\mathbf{W}_Q, \mathbf{W}_K, \mathbf{W}_V \in \mathbb{R}^{d \times d_k}$ are learned projection matrices, and $d_k$ is the dimensionality of the queries and keys. Next, the output is computed using the scaled dot-product attention. The attention output is calculated as follows:
\begin{equation}
\text{Attention}(\mathbf{Q}, \mathbf{K}, \mathbf{V}) = \text{softmax} \left( \frac{\mathbf{Q} \mathbf{K}^T}{\sqrt{d_k}} \right) \mathbf{V}
\end{equation}
In the multi-head attention mechanism, this process is repeated $h$ times, each with different learned projections $\mathbf{W}_Q^{(i)}, \mathbf{W}_K^{(i)}, \mathbf{W}_V^{(i)}$ for each head $h$, resulting in $H$ separate attention outputs. These outputs are concatenated and projected back to the original dimension $d$ using a final learned matrix $\mathbf{W}_O \in \mathbb{R}^{hd_k \times d}$:
\begin{equation}
\text{MultiHead}(\mathbf{Q}, \mathbf{K}, \mathbf{V}) = \text{Concat}(\text{head}_1, \dots, \text{head}_H) \mathbf{W}_O
\end{equation}

where each attention head $\text{head}_h$ is defined as $\text{head}_h = \text{Attention}(\mathbf{Q}^{(h)}, \mathbf{K}^{(h)}, \mathbf{V}^{(h)})$. 
\paragraph{\kvcache} During autoregressive inference, where tokens are generated sequentially, the model has to compute the attention distributions over all previously generated tokens at each step. Without optimisations, this would involve recalculating the key ($\mathbf{K}$) and value ($\mathbf{V}$) projections for every past token at each new step. The \kvcache addresses this inefficiency by storing the key and value projections for each token after they are first computed. Instead of recalculating these projections for past tokens, the model retrieves the cached $\mathbf{K}$ and $\mathbf{V}$ values during subsequent inference steps. 

When generating a new token at time step $t$, the attention computation is performed as:
\begin{equation}
\text{Attention}(\mathbf{Q}_t, [\mathbf{K}_{1:t-1}; \mathbf{K}_t], [\mathbf{V}_{1:t-1}; \mathbf{V}_t])
\end{equation}
where $[;]$ denotes concatenation along the sequence dimension, and $\mathbf{K}_{1:t-1}$ and $\mathbf{V}_{1:t-1}$ are retrieved from memory. The key $\mathbf{K}_t$ and value $\mathbf{V}_t$ for the current token are computed normally.

The \kvcache can significantly reduce computational costs by avoiding redundant calculations. However, storing the cached key and value matrices for every token in the sequence incurs substantial memory usage, which grows linearly with the sequence length. For a model with $L$ layers, $H$ attention heads, and a sequence length of $n$, the total memory required is $L \times H \times n \times d_k \times 2 \times$, 
where the factor of 2 accounts for both the key and value matrices and $\textit{precision}$ represents the number of bytes used to store each value in the memory, typically corresponding to the bit-width of the data type (e.g., 16 bits for half-precision or 32 bits for single-precision floating point).

Though the \kvcache improves the computational efficiency, it requires repeatedly reading potentially large \kvcache from high-bandwidth memory to the streaming multiprocessor during decoding. To address this, recent works~\citep{h2o,FastGen,li2024snapkv,kvcachesurvey2024} have proposed compressing the \kvcache to reduce memory usage.
%While the \kvcache is an effective optimization to enhance inference speed, the substantial memory requirements have prompted exploration into methods for \kvcache compression. 
%
%
%
%
\section{Analysis of the Attention Distributions}
We first examine the attention scores on the language modelling task for a range of popular LLMs.
By analysing the key embeddings and the attention distribution, we observe that key embeddings with low \norm are often associated with higher attention scores.
In~\Cref{fig:attn_norms_kurt}, we provide an example using Llama-2-7b~\citep{touvron2023llama}, where the columns represent different heads, the first row presents the attention distribution over the KV pairs, and the second row presents the \norm of each key embedding. 
We observe that the tokens with high attention scores, such as \verb|"<s>"| and \verb|"."|, have significantly lower \norm values than others.
While \citet{xiao2024efficient} already observed peaked attention distributions for specific tokens, and \citet{darcet2024vision} pointed out the influence of high \norm hidden states on attention maps, we are the first, to the best of our knowledge, to point out the correlation between the \norm of the \emph{key embeddings} and attention score.
Based on our observation, we consider the following research question: can we compress the \kvcache based on the \norm of the key embeddings? 

An intuitive way to estimate the influence of compressing the \kvcache is by examining the attention scores that are dropped due to the compression.
In the following, we formally define this influence.
Given a prompt consisting of $n$ tokens $(x_1, x_2, ..., x_n)$, the LLM first encodes them into a \kvcache --- this step is referred to as the \emph{pre-filling phase}.
Then, the model autoregressively generates the next token $x_{n+1}$. % and uses it as the new input that queries the \kvcache to generate the new next token.
When performing \kvcache compression, some key-value pairs may be dropped and thus cannot be attended to.
We define the attention loss caused by the compression as the sum of the attention scores associated with the dropped KV pairs:
\begin{equation}
\label{eq:attention-loss}
    \mathcal{L}^m_{l,h} = \sum_{p\in D_{l,h}} a_{l,h,p},
\end{equation}
where $a_{l,h,p}$ is the attention score of the $p$-th token in the layer $l$, head $h$.
In \Cref{eq:attention-loss}, $D_{l,h}$ denotes the positions of $m$ pairs of dropped KV, $|D_{l,h}|=m$, which depends on the compression method.
An ideal compression algorithm aims to drop the KV pairs with the lowest attention scores, which will have less impact on the output.
However, such attention scores are unavailable for a compression algorithm since it needs $x_{n+1}$ to query the full \kvcache in advance.
Instead, we drop KV pairs with the highest \norm in key embeddings and use attention loss caused by ideal compression as the reference:
\begin{equation}
    \mathcal{Y}_{l,h}^m = \mathcal{L}_{l,h}^m - \mathcal{L}_{l,h}^{m, ref},
\end{equation}
\noindent where $\mathcal{L}_{l,h}^{m, ref}$ is the reference attention loss, and $\mathcal{Y}_{l,h}^m$ is a non-negative value.
A lower $\mathcal{Y}_{l,h}^m$ indicates a lower difference and thus a higher correlation between the attention score and the \norm.
To measure the overall difference between ideal attention score-based compression and \norm-based compression, we sum up the $\mathcal{Y}_{l,h}^m$ over different numbers of compressed KV pairs:
% $\mathcal{Y}_{l,h} = \sum_{m=1}^{n} \mathcal{Y}_{l,h}^m$.
%
\begin{equation}
\label{eq:norm-attn-diff}
    \mathcal{Y}_{l,h} = \sum_{m=1}^{n} \mathcal{Y}_{l,h}^m.
\end{equation}
We name the $\mathcal{Y}_{l,h}$ as \corrscore, which denotes the attention loss (\Cref{eq:attention-loss}) for a compression method using the ideal attention loss as reference.
\begin{figure}
 \centering
 \begin{subfigure}{.4\textwidth}
     \centering
     \includegraphics[width=\columnwidth]{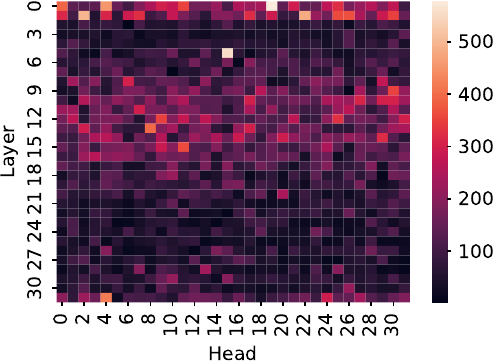}
     \label{fig:norm-attn-diff_llama2}
 \end{subfigure}% 
 \hfill
 \begin{subfigure}{.4\textwidth}
     \centering
     \includegraphics[width=\columnwidth]{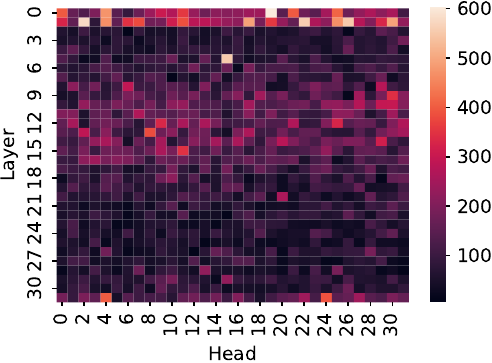}
     \label{fig:norm-attn-diff_llama2_32k}
 \end{subfigure}
 \caption{\corrscore , as defined in \Cref{eq:norm-attn-diff}, for each head and layer in Llama2-7b (left) and Llama2-7b-32k long context model (right). A lower value means a higher correlation between \norm and attention score.}
 \label{fig:norm-attn-diff}
 \end{figure}
%
%
% \begin{figure*}[t]
%     \centering
%     \begin{subfigure}[t]{0.32\textwidth}
%     \includegraphics[width=1\linewidth]{figures/norm_attn_diff/layers_heads_norm_attn_diff_llama-2-7b.pdf}
%     \captionsetup{width=0.9\linewidth}
%     \caption{Llama-2-7b}
%     \end{subfigure}
%     \begin{subfigure}[t]{0.32\textwidth}
%     \includegraphics[width=1\linewidth]{figures/norm_attn_diff/layers_heads_norm_attn_diff.pdf}
%     \captionsetup{width=0.9\linewidth}
%     \caption{Llama-2-7b-80k}
%     \end{subfigure}
%     \begin{subfigure}[t]{0.32\textwidth}
%     \includegraphics[width=1\linewidth]{figures/norm_attn_diff/layers_heads_norm_attn_diff_llama-2-7b-longlora-32k-ft.pdf}
%     \captionsetup{width=0.9\linewidth}
%     \caption{Llama-2-7b-longlora-32k-ft}
%     \end{subfigure}
%     \caption{\corrscore, as defined in \Cref{eq:norm-attn-diff}, of different models. The light square means a high difference between \(L_2\)-Norm. We observe that llama-2 and its fine-tuned long-context versions have a consistent pattern -- the heads in the first two layers and middle layers are more different than others.}
%     \label{fig:norm-attn-diff}
% \end{figure*}
%
%
In \Cref{fig:norm-attn-diff}, we plot the $\mathcal{Y}$ across layers and heads.
We observe that heads in the first two layers and some middle layers around the 12th layer have relatively high $\mathcal{Y}$ values.
The heads in other layers have lower $\mathcal{Y}$ values, indicating a high correlation between \norm and attention score.

By leveraging this correlation, we can compress the \kvcache based on the \norm of key embeddings. Optionally, we can skip the compression at the layers with low correlation. We show ablation experiments skipping layers in \Cref{sec:lm_more_results}. 
\section{Experiments}
\label{sec:experiments}
\begin{figure*}[t]
    \centering
        \begin{subfigure}{0.325\textwidth}
            \centering
            \includegraphics[width=\textwidth]{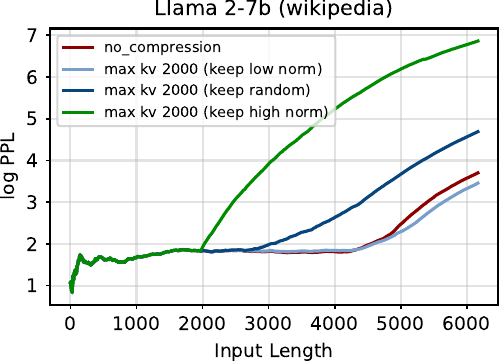}
            %\caption{\deitt on Food101}
        \end{subfigure} 
        \begin{subfigure}{0.325\textwidth}
            \centering
            \includegraphics[width=\textwidth]{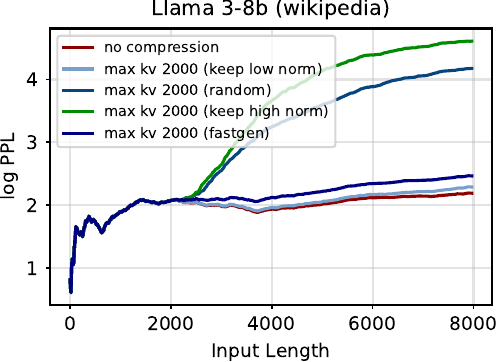}
            %\caption{\deitt on Cifar100}
        \end{subfigure} 
        \begin{subfigure}{0.325\textwidth}
            \centering
            \includegraphics[width=\textwidth]{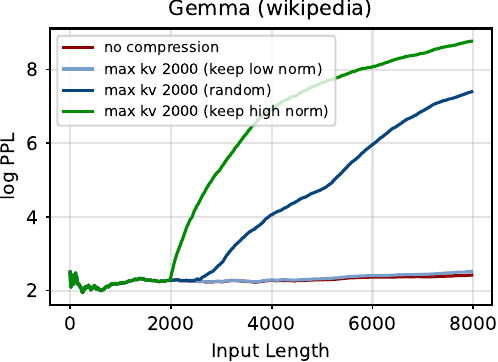}
            %\caption{\deitt on Flowers102}
        \end{subfigure} \\
    \caption{Perplexity for Llama 2-7b, Llama 3-8b and Gemma on language modelling task on wikipedia  dataset.Additional results on coding dataset are available in \Cref{sec:lm_more_results} 
    \label{fig:language_modelling_results}
    }
\end{figure*}

We evaluate our method on language modelling and two long-context modelling tasks, i.e., needle-in-a-haystack and passkey retrieval. In addition, we test on tasks from LongBench~\citep{longbench-zhang-etal-2024}, specifically devised to evaluate the model's long context abilities.
%
% In all cases, we use the \norm-based compression, but we also tested compression based on sparsity with similar results for language modeling and worse for KV compression. \zhaoyu{I think it is better to put sparsity analysis in the appendix (including visualisation and language modelling experiments). 1. we do not investigate this too much; 2. the current sparsity-based method performs badly on needle and passkey; 3. this paper is just about \norm-based compression; 4. it does not directly relate to the norm-based, which makes readers confusing} 
%
Based on the observation supported by \Cref{fig:norm-attn-diff}, the heads in the first two layers usually have a low correlation between \norm and attention score, so we do not perform compression on these layers as default.
We conduct experiments to investigate the impact of compression on different layers in \Cref{sec:lm_more_results}.
\paragraph{Language Modelling Tasks}
For language modelling, we let the \kvcache grow until a specific pre-defined length and subsequently start to discard the tokens with the highest \norm.
We show in \Cref{fig:language_modelling_results} that evicting even up to the 50\% of \kvcache does not impact perplexity. Perplexity increases, as expected, once we exceed the pre-training context length. We show more results, including next token accuracy in \Cref{sec:lm_more_results}.  
To further verify that keys with low \norm capture significant information, we test other eviction strategies, i.e. keeping tokens with highest \norm and keeping random tokens. It is clear from \Cref{fig:language_modelling_results} that discarding tokens with low \(L_2\) impairs performance, even more so than random discarding, thus highlighting the importance of these low \norm keys.

\begin{figure}[t]
\centering
\begin{subfigure}{.43\textwidth}
    \centering
    \includegraphics[width=\columnwidth]{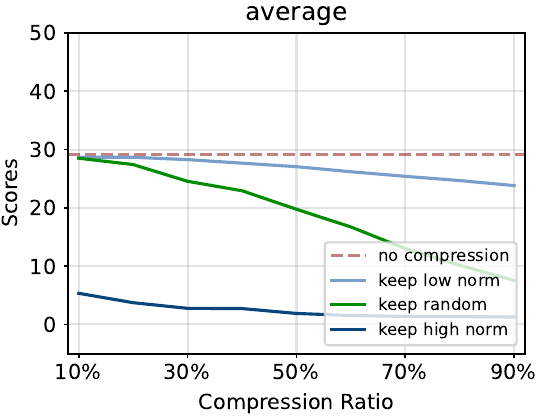}
    %\caption{Overall accuracy of llama-2-7b-80k on the needle-in-a-haystack task.  More results in \Cref{sec:more-long-context-tasks-results}.}
    %\label{fig:needle-in-a-haystack-overall}
\end{subfigure}% 
\hfill
\begin{subfigure}{.43\textwidth}
    \centering
    \includegraphics[width=\columnwidth]{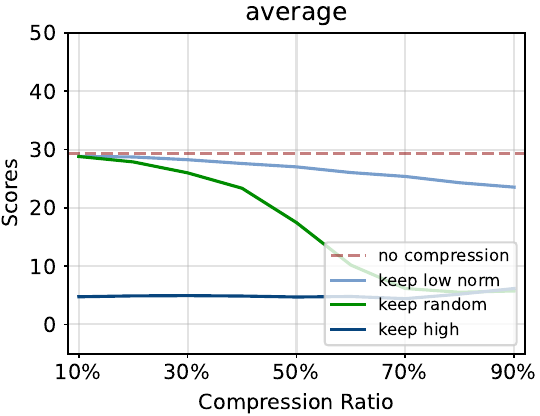}
    % \caption{Overall accuracy of llama-2-7b-80k on the passkey retrieval task.  More results in \Cref{sec:more-long-context-tasks-results}.}
    % \label{fig:longbenvch-avergage}
\end{subfigure}
\caption{Overall scores on LongBench \citep{longbench-zhang-etal-2024} of Llama3.1-8b (left) and llama-2-7b-80k (right) for different compression ratios ranging from $0\%$ to $90\%$. }
\label{fig:longbench_average}
\end{figure}

\begin{figure}[t]
\centering
\begin{subfigure}{.43\textwidth}
    \centering
    \includegraphics[width=\columnwidth]{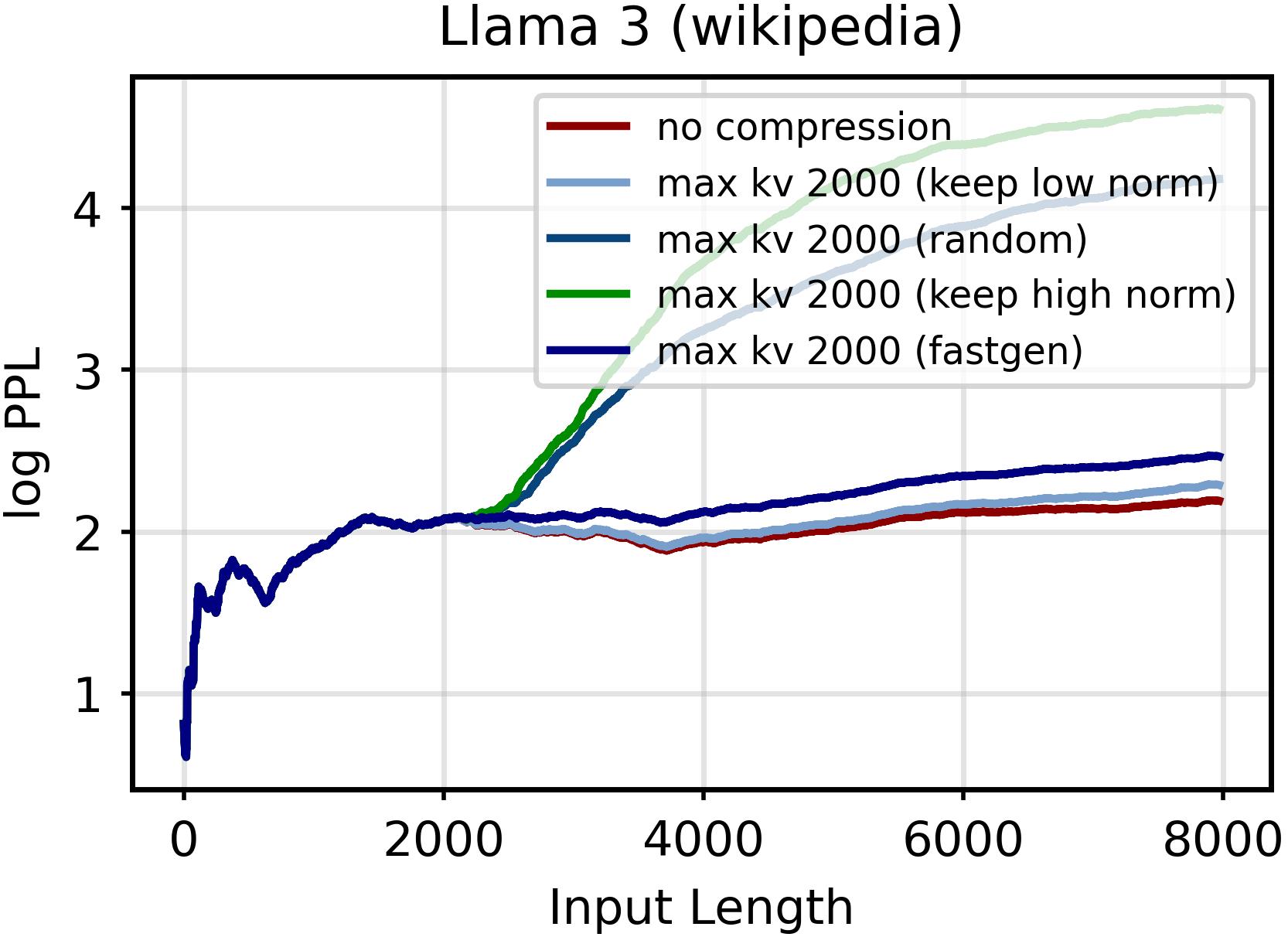}
\end{subfigure}% 
\hfill
\begin{subfigure}{.43\textwidth}
    \centering
    \includegraphics[width=\columnwidth]{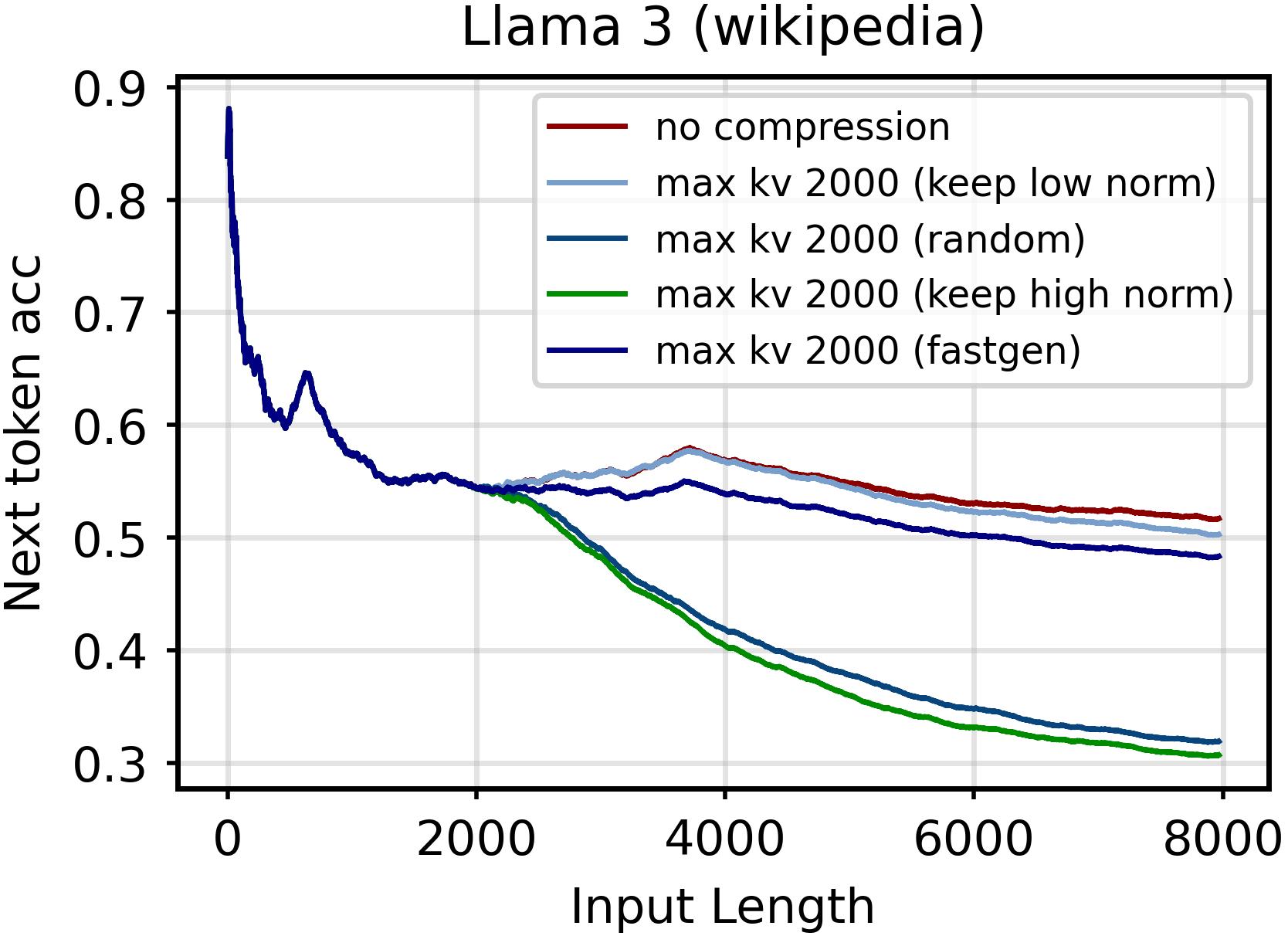}
\end{subfigure}
\caption{Perplexity and next token accuracy of Llama3-8b on the wikipedia dataset when compared to FastGen \citep{FastGen} (only local, special and punctuation tokens).}
\label{fig:fastgen_results}
\end{figure}

% \subsection{Long-Context Modelling Tasks}
\paragraph{Pressure Test on Long-Context Tasks}
\begin{figure}[b]
\centering
\begin{subfigure}{.43\textwidth}
    \centering
    \includegraphics[width=\columnwidth]{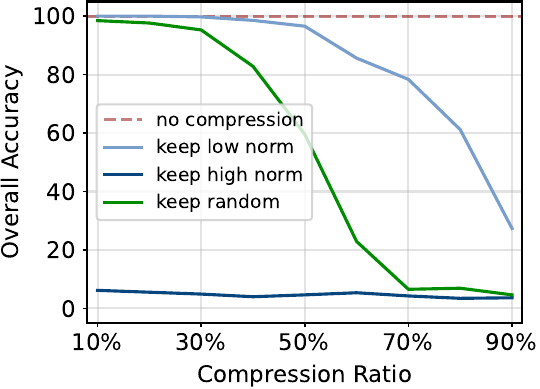}
    %\caption{Overall accuracy of llama-2-7b-80k on the needle-in-a-haystack task.  More results in \Cref{sec:more-long-context-tasks-results}.}
    \caption{\footnotesize Accuracy on the needle-in-a-haystack task.}
    \label{fig:needle-in-a-haystack-overall}
\end{subfigure}% 
\hfill
\begin{subfigure}{.43\textwidth}
    \centering
    \includegraphics[width=\columnwidth]{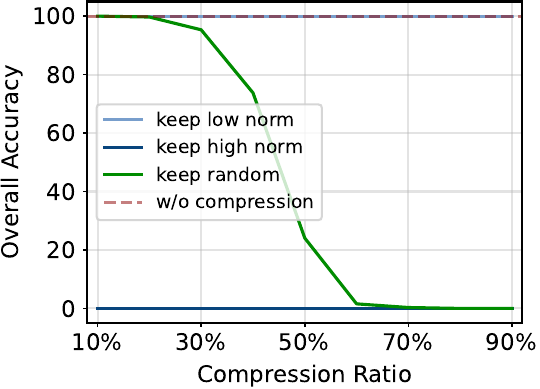}
    % \caption{Overall accuracy of llama-2-7b-80k on the passkey retrieval task.  More results in \Cref{sec:more-long-context-tasks-results}.}
    \caption{\footnotesize Accuracy on the passkey retrieval task.}
    \label{fig:passkey-retrieval-overall}
\end{subfigure}
%\caption{\corrscore , as defined in \Cref{eq:norm-attn-diff}, for each head and layer in Llama2-7b (left) and Llama2-7b-32k long context model (right). A lower value means a higher correlation between \norm and attention score.}
\caption{Overall accuracy of llama-2-7b-80k on the needle-in-a-haystack task  passkey retrieval task.}
\label{fig:llama2-80k-long-context}
\end{figure}
The needle-in-a-haystack task~\citep{needleinhaystack} and passkey retrieval task~\citep{passkey-retrieval} are two synthetic tasks that are widely used to pressure test the long-context modelling capability of LLMs.
% We evaluate the compression capability on two long context-modelling tasks, needle-in-a-haystack~\citep{needleinhaystack} and passkey retrieval~\citep{passkey-retrieval}.
%
In both tasks, the model needs to identify and retrieve the important information from a long context to generate correct answers.
Thus, these tasks test the compression method's ability to keep important KV pairs and drop redundant ones.
In \Cref{fig:needle-in-a-haystack-overall} and \Cref{fig:passkey-retrieval-overall}, we present the experimental results of Llama-2-7b-80k~\citep{yaofu-data-eng}. We analyse additional models in \Cref{sec:more-long-context-tasks-results}.
As shown in \Cref{fig:needle-in-a-haystack-overall}, the model can preserve its performance on the needle-in-a-haystack task while compressing 30\% of the \kvcache, and maintain 99\% accuracy when compressing 50\% of the \kvcache. Additionally, the model can achieve 100\% accuracy on the passkey retrieval task even when compressing 90\% of the \kvcache, as shown in \Cref{fig:passkey-retrieval-overall}.
Moreover, we compare other eviction strategies, like keeping KV pairs with low \norm, keeping KV pairs with high \norm, and keeping random KV pairs.
In \Cref{fig:needle-in-a-haystack-overall} and \Cref{fig:passkey-retrieval-overall}, we observe that the model cannot answer correctly when keeping only high \norm KV pairs, obtaining near zero and zero accuracy on the needle-in-a-haystack and passkey retrieval tasks, respectively.
When we randomly compress the \kvcache, the performance decreases significantly faster than keeping low \norm KV pairs. 
The above analysis indicates that KV pairs with low \norm are critical to generating the correct answer and thus  contain important information. 
%
% In \Cref{sec:more-long-context-tasks-results} we show detailed results on long context modelling tasks.
%
%
\paragraph{Experiments on LongBench}
Additionally, we evaluate on LongBench \citep{longbench-zhang-etal-2024}. We test on several subsets, including NarrativeQA~\citep{narrativeqa}, Qasper~\citep{qasper}, HotpotQA~\citep{hotpotqa}, 2WikiMQA~\citep{wikimqa}, and QMSum~\citep{qmsum}. We report the results for the recently released long context Llama3.1 and Llama 2-7b 80k in \Cref{fig:longbench_average}. In addition, we show the complete per-subset results in \Cref{sec:more-long-context-tasks-results}.
The experimental results show that compressing the \kvcache with low \norm only introduces a small accuracy decrease even when compressing 50\% \kvcache, while compressing \kvcache with high \norm results in almost zero accuracy.
%
%
%
%
% \subsection{Comparison with FastGen}
\paragraph{Comparison with FastGen}
We use FastGen~\citep{FastGen}, a popular method for \kvcache compression, as a baseline for assessing the effectiveness of our method. It is important to note that, like the majority of methods in the literature, FastGen utilises attention scores, which makes it incompatible with the popular FlashAttention~\citep{dao2022flashattention}, thereby limiting its efficiency and usability. For a fair comparison, we implement FastGen without using the attention scores, i.e., we only consider local, punctuation and special tokens. We perform experiments on language modelling with the Llama3 model~\citep{llama3}. Our method still outperforms FastGen with up to 50\% \kvcache eviction. We show the results in \Cref{fig:fastgen_results}.
%
% \begin{figure}[t]
%     \centering
%     \subfloat[]{\includegraphics[width=0.43\textwidth]{figures/fastgen/Fastgen_Llama_3_wikipedia_ppl.jpeg}}\hfill
%     \subfloat[]{\includegraphics[width=0.43\textwidth]{figures/fastgen/Fastgen_Llama_3_wikipedia_acc.jpeg}}\\
%     \caption{Perplexity and next token accuracy of Llama3-8b on the wikipedia dataset when compared to FastGen \citep{FastGen} (only local, special and punctuation tokens).}
%     \label{fig:fastgen_results}
% \end{figure}

\section{Analysis}
\begin{figure}[t]
    \centering
    \begin{subfigure}[t]{0.43\textwidth}
    \includegraphics[width=1\linewidth]{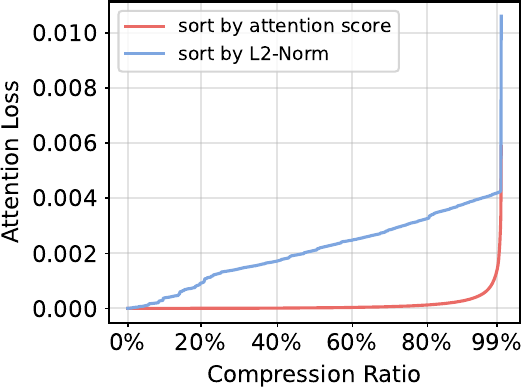}
    \captionsetup{width=0.9\linewidth}
    \caption{Layer-7 Head-10, high correlation between attention score and \(L_2\)-Norm.}
    \label{fig:fig:norm-attn-diff-layer7-head10}
    \end{subfigure}
    \hfill
    \begin{subfigure}[t]{0.43\textwidth}
    \includegraphics[width=1\linewidth]{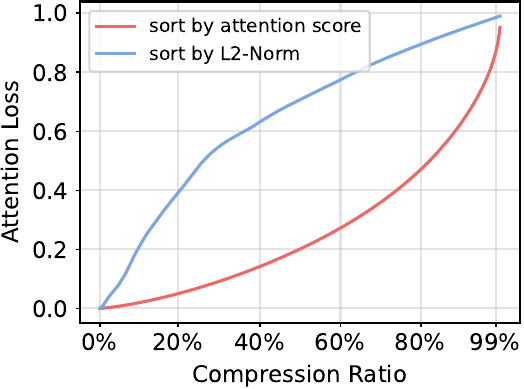}
    \captionsetup{width=0.9\linewidth}
    \caption{Layer-0 Head-0, low correlation between attention score and \(L_2\)-Norm. }
    \label{fig:norm-attn-diff-layer0-head0}
    \end{subfigure}
    \caption{
    Attention loss of ideal compression and \norm-based compression in Llama-2-7b-80k.
    The $x$-axis represents the compression ratio; the $y$-axis represents the attention loss (defined by \Cref{eq:attention-loss}) The results average over 1024 chunks on Wikipedia, with a length of 1024.
    }
    \label{fig:norm-attn-diff-layer-head}
\end{figure}

\begin{figure}[b]
    \centering
    \subfloat{\includegraphics[width=0.3\textwidth]{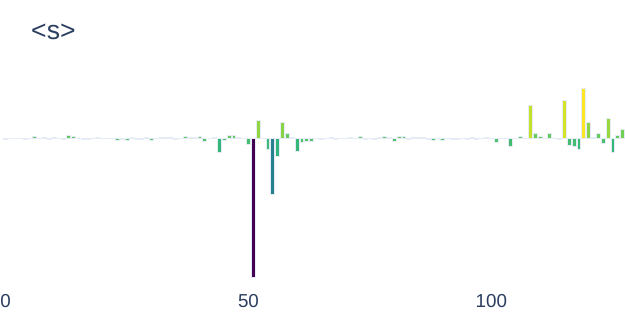}}\hfill
    \subfloat{\includegraphics[width=0.3\textwidth]{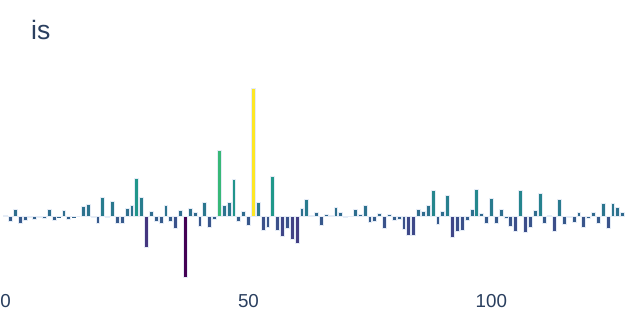}} \hfill
    \subfloat{\includegraphics[width=0.3\textwidth]{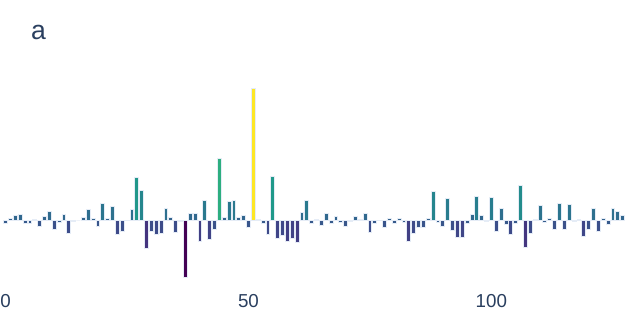}} \\
    \subfloat{\includegraphics[width=0.3\textwidth]{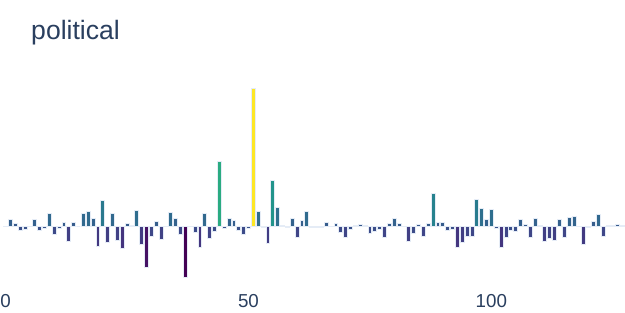}} \hfill
    \subfloat{\includegraphics[width=0.3\textwidth]{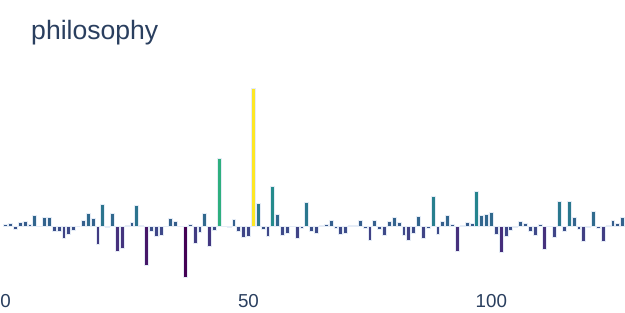}} \hfill
    \subfloat{\includegraphics[width=0.3\textwidth]{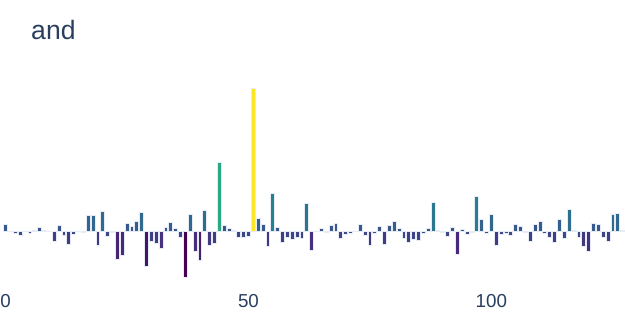}} \\
    \caption{Key projections of the bos token $<s>$ vs other tokens. Each value represents the activation in a specific dimension for the embedding of the key projection. We found similar patterns across almost all heads and layers and in multiple texts. Only a few peaked activations ($\sim 50$, $\sim 56$ and $ \sim 120$) control the attention mechanism (see \Cref{fig:attentions}). More plots like this in  \Cref{sec:more_token_emb}}
    \label{fig:sparse_projections}
\end{figure}

\begin{figure}[t]
    \centering
    \subfloat{\includegraphics[width=0.7\textwidth]{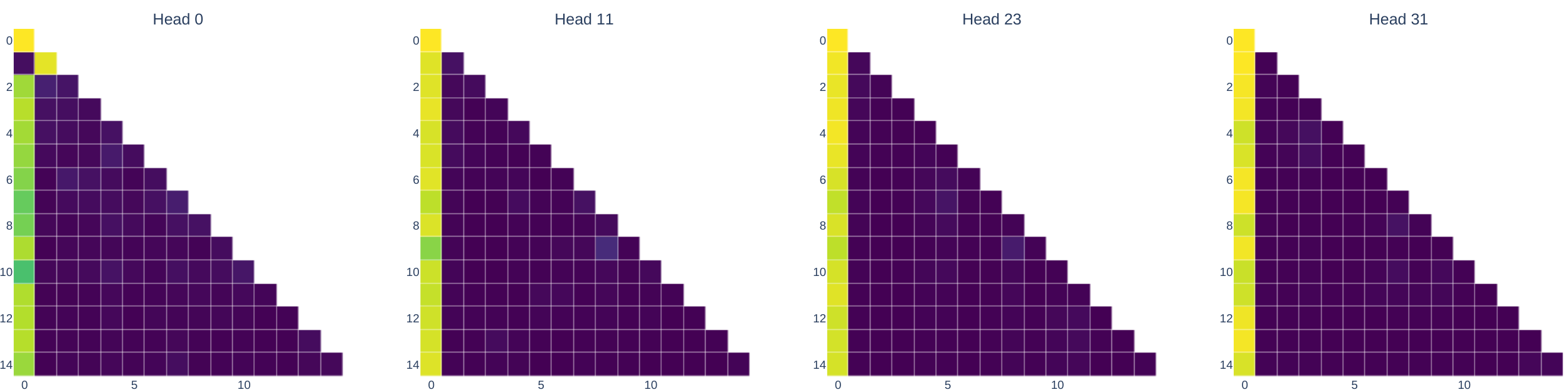}} \\
    \subfloat{\includegraphics[width=0.7\textwidth]{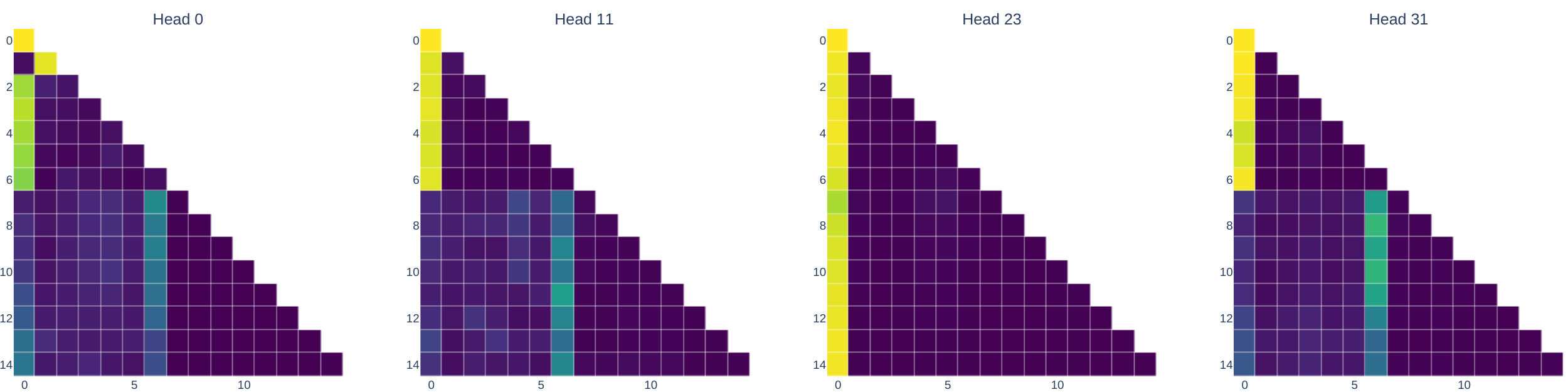}}
    \caption{How the attention maps change if we set to zero a random activation (top) vs the specific peaked activations in the keys (bottom). We are setting the values at iteration 5.}
    \label{fig:attentions}
\end{figure}

\paragraph{Attention score loss when using $L_2$ norm}
We discuss further the correlation between \norm and attention scores.
We already displayed in \Cref{fig:norm-attn-diff} the \norm and attention correlation across heads and layers using the original Llama2-7b and the long context Llama2-7b-32k and Llama2-7b-80k. We can see that patterns are quite consistent across all the models. To better visualise how correlation varies across different heads,  in \Cref{fig:norm-attn-diff-layer-head}, we only consider two heads from layer 10 and layer 0 and show the \corrscore from \Cref{eq:attention-loss}. As expected, we see that in layer 0, the difference is larger due to a lower correlation. 
% This implies that compressing based on \norm can be less effective depending on the layer and head considered. 
%
%
\paragraph{Relationship between embedding and $L_2$ norm}
So far, we have identified a correlation between the \norm of token key embeddings and the corresponding attention scores. This observation, while primarily empirical, it offers a direction for further explorations. 
Our investigation into the distribution of key embeddings revealed that tokens with lower \norm tend to exhibit \emph{sparse activations} with only a few dimensions showing significantly high values, while the majority of the dimensions remain near zero. This pattern suggests that the embeddings of these tokens are not fully utilising the available vector space, focusing their activations on a narrow subset of dimensions. \Cref{fig:sparse_projections} illustrates several examples of such tokens, highlighting the difference between tokens with high and low \norm.
Interestingly, this sparsity aligns with the concept of "sink" tokens, as identified in previous studies \citep{xiao2024efficient}. These tokens capture a direction in the embedding space such that many queries align closely with it, leading to increased attention scores for these tokens.  Specifically, when the key embeddings of certain tokens are dominated by a limited set of dimensions, they create a focal point, attracting a wide range of queries -- regardless of their individual content -- and amplifying their attention weights.
We hypothesise that the lower \norm reflects a partial use of the available embedding space, leading to increased attention for these tokens. To examine this hypothesis, we zeroed out the dimensions responsible for the peaked activations in low-norm key embeddings and observed significant changes in attention maps (\Cref{fig:attentions}).  In contrast, altering random dimensions did not produce the same effect, highlighting the importance of these specific dimensions. This finding suggests that the \norm may serve as a proxy for the extent to which an embedding utilises the available vector space and, consequently, the degree to which it influences attention. Lower \norm appears to correspond to embeddings that drive disproportionately high attention values due to their alignment with a common direction. 
\citet{cancedda-2024-spectral} offers additional insight into this phenomenon, suggesting that attention sinks engage with other tokens through a “dark” subspace within the embedding space.
\section{Related Work}
Recently, various long-context LLMs, such as Gemini-Pro-1.5~\citep{reid2024gemini}, Claude-3~\citep{anthropic2024claude}, and GPT4~\citep{achiam2023gpt}, have shown the promising capability to process hundred thousands of tokens in the context.
The increased number of input lengths results in a high decoding latency; thus, there has been a growing interest in speeding up the decoding with long contexts.
Some works propose efficient memory management strategies to reduce the IO time overheads, e.g., PageAttention~\citep{vllm}, Infinite-LLM~\citep{infinitellm} and vAttention~\citep{vattention}.
Another line of research focuses on compressing the \kvcache to improve efficiency.
DMC~\citep{nawrot2024dynamic} compresses \kvcache by dynamically merging tokens while requiring expensive continual pre-training.
For fine-tuning free compression strategy, H2O~\citep{h2o} identifies important KV pairs by leveraging the attention scores from all queries, FastGen~\citep{FastGen} leverages the different attention patterns in different heads for compression, and SnapKV~\citep{li2024snapkv} selects KV pairs based on attention scores from user's query. 
%We refer the reader to ~\citet{kvcachesurvey2024} for a broader survey on this topic.
%
Unlike these works, our method only utilises the \norm of embedding for compression \textit{without leveraging the attention information}, and to the best of our knowledge, we are the first to find that the influence of a KV pair can be determined by \norm. 
Previous work~\citep{darcet2024vision} finds the hidden states with high \norm usually aggregate more important and global information.
On the other hand, our findings indicate that a low \norm of key embedding generally results in a high attention score.
Concurrently to this work, \citet{guo2024attentionscoreneedtoken} uses the $L_1$ norm of values in the \kvcache and attention scores for compression. 
\section{Conclusions}
In this paper, we introduced a simple yet highly effective strategy for \kvcache compression in LLMs based on the \norm of key embeddings.
We show that there is a significant correlation between the \norm of a key embedding and its attention score.
Leveraging this observation, we compress the \kvcache by retaining only those keys with the lowest \norm.
Our experimental results on various tasks show that our compression strategy maintains the predictive accuracy of the model while significantly reducing the memory footprint.
Our approach is straightforward and can be applied directly to any transformer-based, decoder-only LLM.
\section{Limitations}
\label{sec:limitations}
%
% While our research provides valuable insights, our experiments and analyses were conducted exclusively using relatively small-size models (models from Llama's family and Gemma up to 8 billion parameters). In future work, we will evaluate our proposed method on larger-scale models to verify that our findings generalise across models and architectures.
While our research offers valuable insights, we tested only on relatively small models (Llama family and Gemma up to 8 billion parameters). In future work, we will assess our method on larger-scale models to ensure our findings generalize
Additionally, while we show that the \norm played a significant role in our experiments, we do not have a comprehensive theoretical explanation for why this is the case. Understanding the underlying reasons behind the importance of the \norm would require further theoretical exploration and empirical validation. 
Finally, we observed (\Cref{fig:norm-attn-diff}) that compressing based on \norm can be less effective depending on the layer and head considered, and we intend to investigate per-head compression ratios to leverage this observation. 
\section{Acknowledgments}
Alessio Devoto was supported by Sapienza Grant RM1221816BD028D6 (DeSMOS).
Yu Zhao was partly supported by the UKRI Centre for Doctoral Training in Natural Language Processing, funded by UK Research and Innovation (grant EP/S022481/1) and the University of Edinburgh, School of Informatics.
Pasquale Minervini was partially funded by ELIAI (The Edinburgh Laboratory for Integrated Artificial Intelligence), EPSRC (grant no.\ EP/W002876/1), an industry grant from Cisco, and a donation from Accenture LLP.
This work was supported by the Edinburgh International Data Facility (EIDF) and the Data-Driven Innovation Programme at the University of Edinburgh.
\bibliographystyle{plainnat}
\bibliography{references}

\clearpage
\appendix
\section{More results on Language modelling task}
\label{sec:lm_more_results}

\begin{figure*}[t]
    \centering
    \begin{subfigure}[t]{0.48\textwidth}
    \includegraphics[width=1\linewidth]{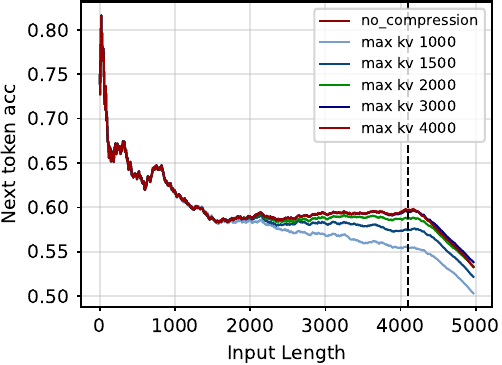}
    \end{subfigure} \hfil
    \begin{subfigure}[t]{0.48\textwidth}
    \includegraphics[width=1\linewidth]{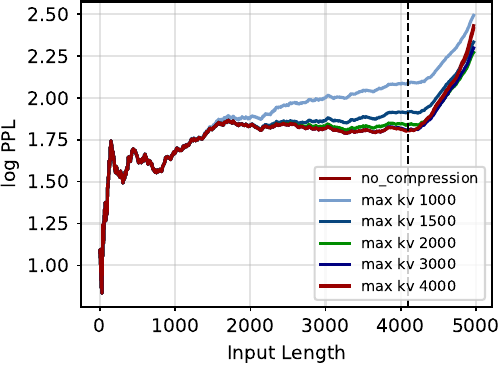}
    \end{subfigure}
    \caption{Results on language modelling task when skipping the first layer.}
    \label{fig:lm_ppl_acc_abl0}
    
    \medskip
    \begin{subfigure}[t]{0.48\textwidth}
    \includegraphics[width=1\linewidth]{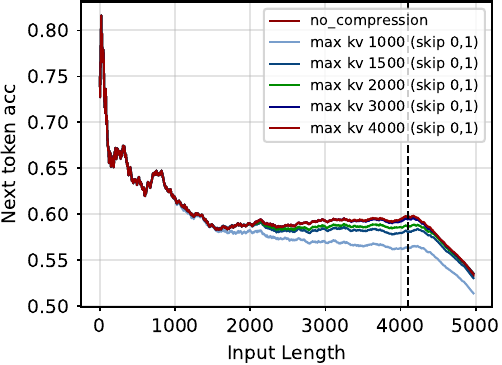}
    \end{subfigure}\hfil
    \begin{subfigure}[t]{0.48\textwidth}
    \includegraphics[width=1\linewidth]{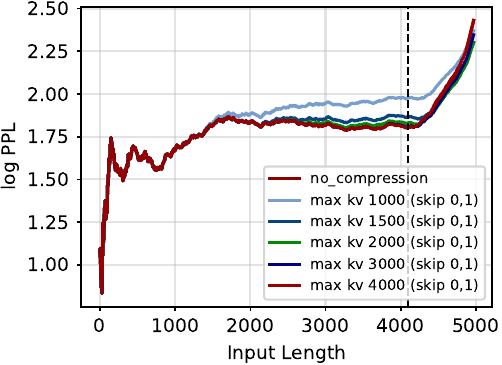}
    \end{subfigure}
    \caption{Results on language modelling task when skipping the first two layers.}
    \label{fig:lm_ppl_acc_abl12}

    \medskip
    \begin{subfigure}[t]{0.48\textwidth}
    \includegraphics[width=1\linewidth]{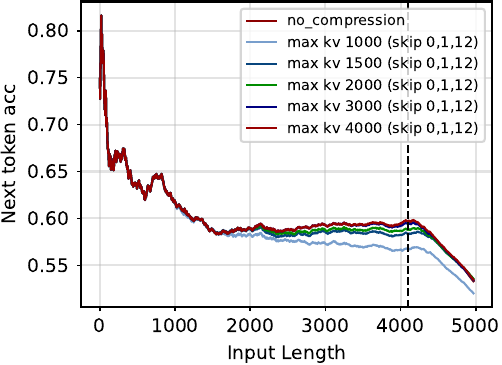}
    \end{subfigure}\hfil
    \begin{subfigure}[t]{0.48\textwidth}
    \includegraphics[width=1\linewidth]{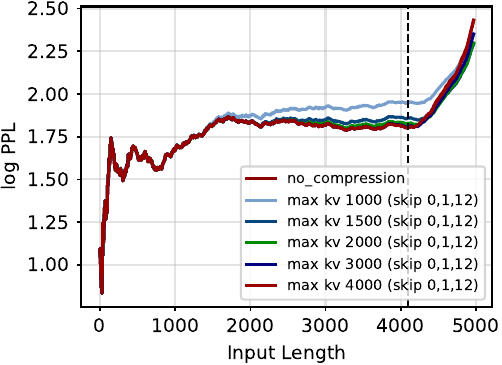}
    \end{subfigure}
    \caption{Results on language modelling task when skipping layers 0,1 and 12.}
    \label{fig:lm_ppl_acc_abl1212}
\caption{Skipping compression at different layers with Llama2-7b}
\label{fig:lm_overall_results}
\end{figure*}

In the following, we show results when performing compression only on layers that show a lower correlation between \norm and attention score. We show in \cref{fig:lm_overall_results} that for language modelling tasks, the different layer drop has little impact on final accuracy and perplexity. The difference becomes significant only when the \kvcache is pruned to retain only one thousand pairs. All experiments are averaged over 50 chunks from English Wikipedia.

\begin{figure}[t]
    \centering
     \begin{subfigure}{0.325\textwidth}
            \centering
            \includegraphics[width=\textwidth]{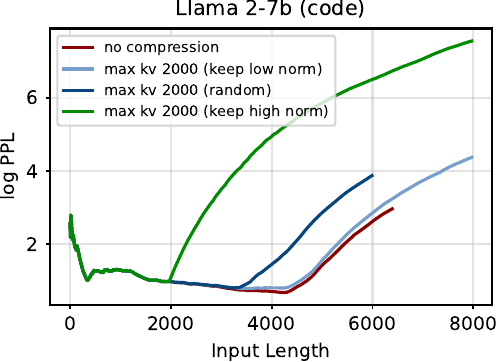}
        \end{subfigure} 
        \begin{subfigure}{0.325\textwidth}
            \centering
            \includegraphics[width=\textwidth]{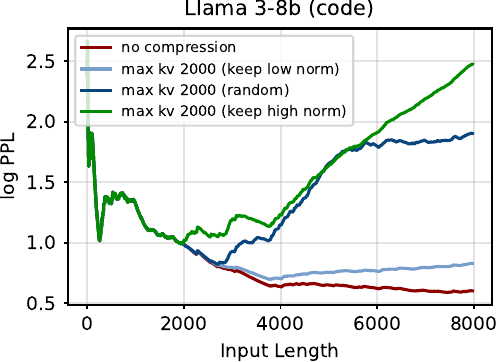}
        \end{subfigure} 
        \begin{subfigure}{0.33\textwidth}
            \centering
            \includegraphics[width=\textwidth]{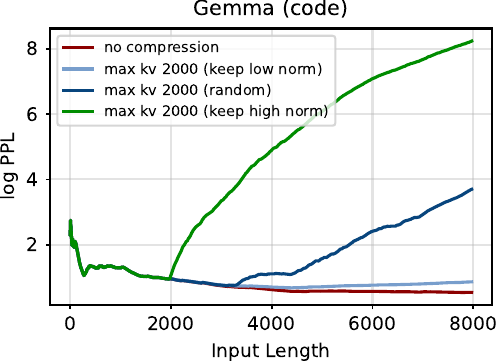}

        \end{subfigure}
\end{figure}
\section{More Results on Long-Context Modelling Tasks}
\label{sec:more-long-context-tasks-results}
\begin{figure*}[!ht]
    \centering
    \begin{subfigure}[t]{0.48\textwidth}
    \includegraphics[width=1\linewidth]{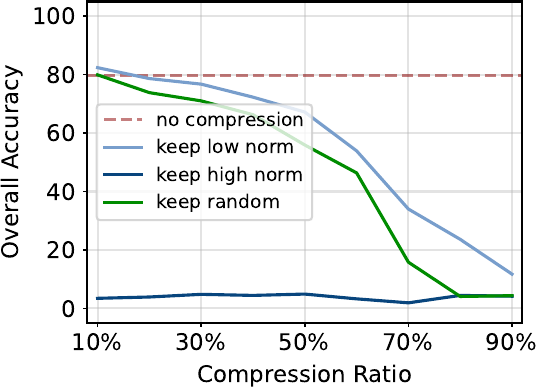}
    \captionsetup{width=0.9\linewidth}
    \caption{Overall accuracy of Llama-2-7b-longlora-32k-ft on the needle-in-a-haystack task.}
    \label{fig:llama-2-7b-longlora-32k-ft-needle}
    \end{subfigure}
    \begin{subfigure}[t]{0.48\textwidth}
    \includegraphics[width=1\linewidth]{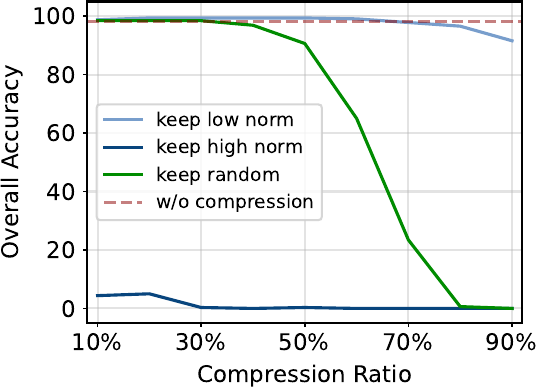}
    \captionsetup{width=0.9\linewidth}
    \caption{Overall accuracy of Llama-2-7b-longlora-32k-ft on the passkey retrieval task.}
    \label{fig:llama-2-7b-longlora-32k-ft-passkey}
    \end{subfigure}
    \caption{Evaluation results of Llama-2-7b-longlora-32k-ft on the needle-in-a-haystack and passkey retrieval tasks.}
    \label{fig:llama-2-7b-longlora-32k-ft-long-context-results}
\end{figure*}

In addition to llama-2-7b-80k~\citep{yaofu-data-eng}, we test the compression method using llama-2-7b-longlora-32k-ft~\citep{longlora} on the needle-in-a-haystack and passkey retrieval tasks.
As shown in~\cref{fig:llama-2-7b-longlora-32k-ft-needle}, we can see that compressing 30\% of \kvcache only results in a slight performance degradation on the needle-in-a-haystack task.
We also observe that the performance even increases slightly when we compress 10\% of \kvcache.
In figure~\cref{fig:llama-2-7b-longlora-32k-ft-passkey}, we observe that the llama-2-7b-longlora-32k-ft maintains 100\% performance when compressing 80\% of \kvcache and only as a slight decrease when compressing 90\% of \kvcache.
Furthermore, the model fails to generate correct answers if we compress KV pairs with low \norm and keep high \norm ones.
The evaluation results of llama-2-7b-longlora-32k-ft are consistent with the llama-2-7b-80k, which further indicates the effectiveness of compressing \kvcache using \norm.

\subsection{Analysis of Skipped Layers}
\begin{figure*}[!ht]
    \centering
    
    \begin{subfigure}[t]{0.48\textwidth}
    \includegraphics[width=1\linewidth]{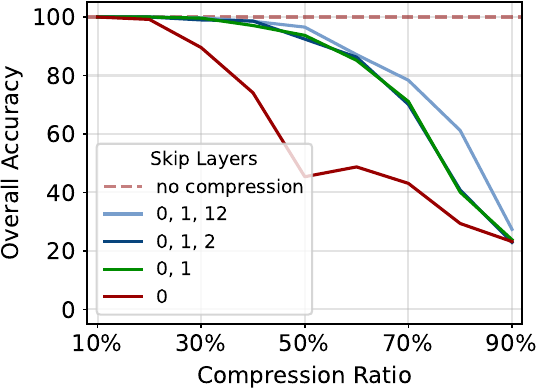}
    \captionsetup{width=0.9\linewidth}
    \caption{Overall accuracy of Llama-2-7b-80k on the needle-in-a-haystack task.}
    \label{fig:llama-2-7b-80k-needle-skip-layers}
    \end{subfigure}
    \begin{subfigure}[t]{0.48\textwidth}
    \includegraphics[width=1\linewidth]{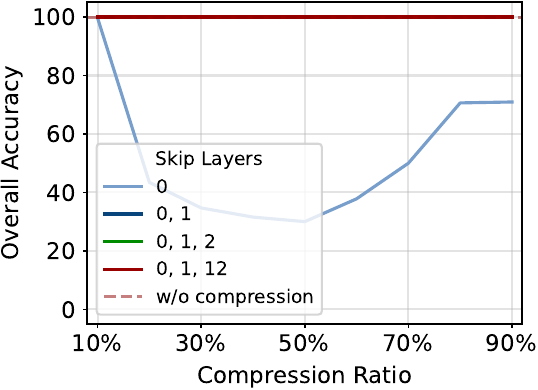}
    \captionsetup{width=0.9\linewidth}
    \caption{Overall accuracy of Llama-2-7b-80k on the passkey retrieval task.}
    \label{fig:llama-2-7b-80k-passkey-skip-layers}
    \end{subfigure}

    \begin{subfigure}[t]{0.48\textwidth}
    \includegraphics[width=1\linewidth]{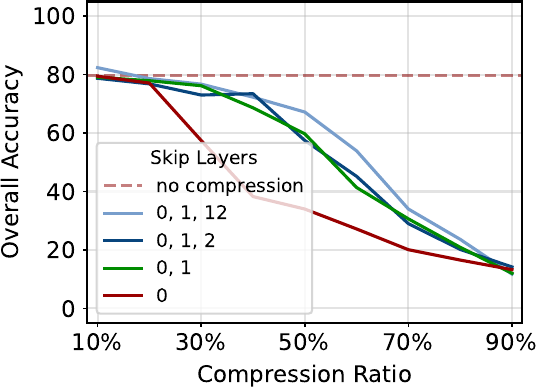}
    \captionsetup{width=\linewidth}
    \caption{Overall accuracy of Llama-2-7b-longlora-32k-ft on the needle-in-a-haystack task.}
    \label{fig:llama-2-7b-longlora-32k-ft-needle-skip-layers}
    \end{subfigure}
    \begin{subfigure}[t]{0.48\textwidth}
    \includegraphics[width=1\linewidth]{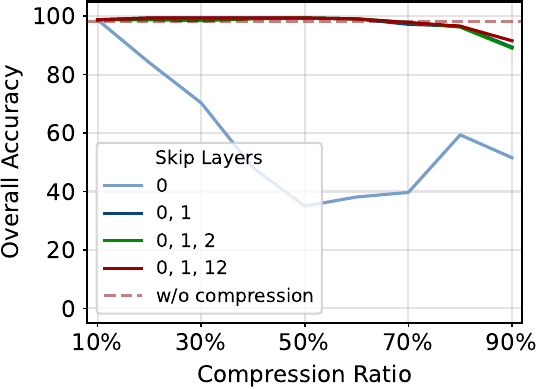}
    \captionsetup{width=0.9\linewidth}
    \caption{Overall accuracy of Llama-2-7b-longlora-32k-ft on the passkey retrieval task.}
    \label{fig:llama-2-7b-longlora-32k-ft-passkey-skip-layers}
    \end{subfigure}
    \caption{Analysing of skipping different layers for compression.}
    \label{fig:skip-different-layers}
\end{figure*}

As shown in~\cref{fig:norm-attn-diff}, we find heads in the first two layers and the middle layers have a relatively low correlation between attention scores and \norm.
Thus, we conduct experiments to analyse the impact of skipping layers that have a low correlation for compression.
As shown in~\cref{fig:llama-2-7b-80k-needle-skip-layers} and~\cref{fig:llama-2-7b-longlora-32k-ft-needle-skip-layers}, we observe that only skipping the first layer (layer-0) decreases the performance on the needle-in-a-haystack task significantly.
We can see that skipping the first two layers (layer-0,1) has a similar performance compared to skipping the first three layers (layer-0,1,2).
%
% These results suggest the effectiveness of analysing correlation using \corrscore.
%
Furthermore, as shown in~\cref{fig:llama-2-7b-80k-passkey-skip-layers} and~\cref{fig:llama-2-7b-longlora-32k-ft-passkey-skip-layers}, only skipping the first layer can result in significant performance degradation.
We also find that the compression ratio is not proportional to the overall accuracy of models in the passkey retrieval task when we compress the first layer, where the accuracy shows a U-shape curve regarding the compression ratio.

% \begin{figure*}[!ht]
%     \centering

%     \caption{Passkey retrieval}
%     \label{fig:passkey-retrieval}
% \end{figure*}

\begin{figure*}
    \centering
    \begin{subfigure}[t]{0.95\textwidth}
    \includegraphics[width=1\linewidth]{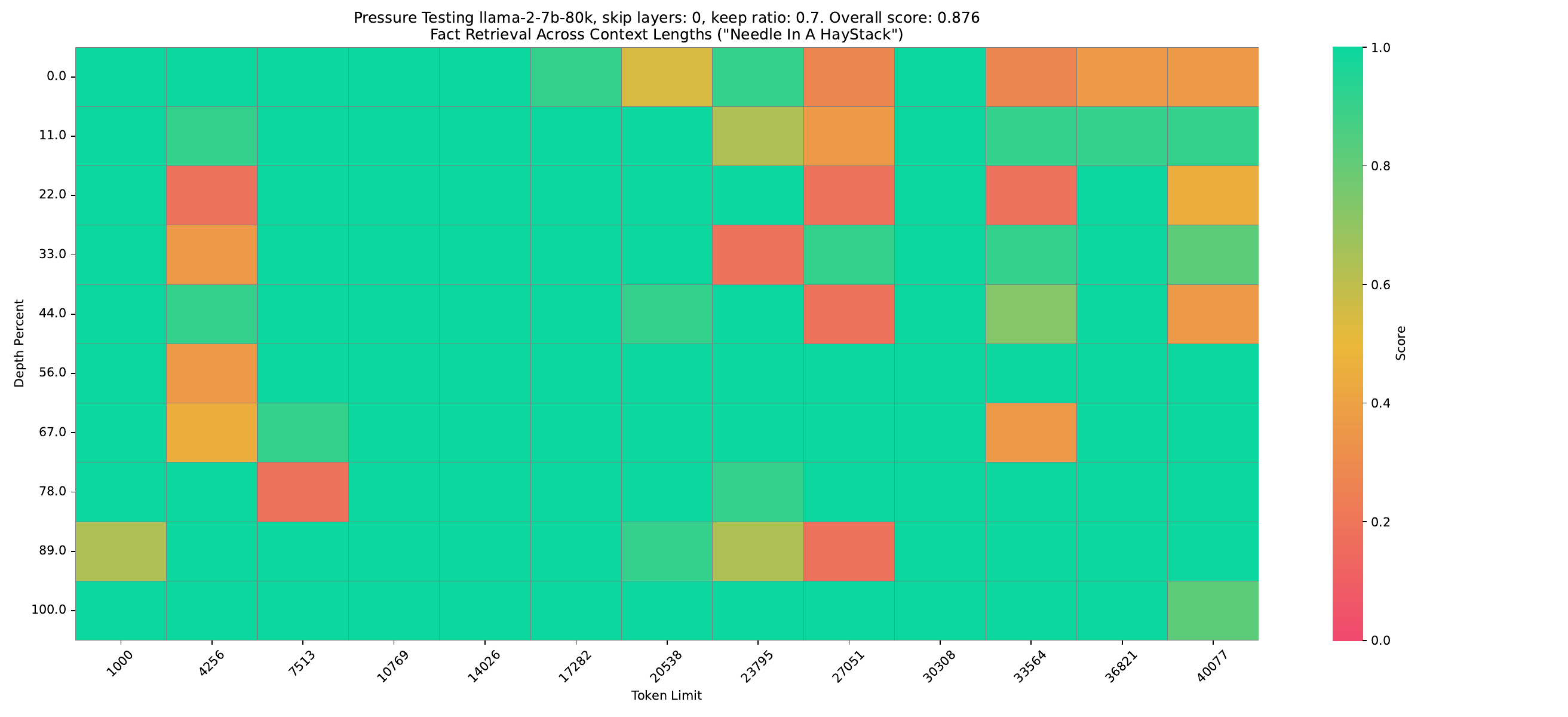}
    \captionsetup{width=0.9\linewidth}
    \caption{Llama-2-7b-80k, skip layer-0, compression ratio 30\%}
    % \label{fig:norm-attn-diff-llama-2-7b}
    \end{subfigure}

    \centering
    \begin{subfigure}[t]{0.95\textwidth}
    \includegraphics[width=1\linewidth]{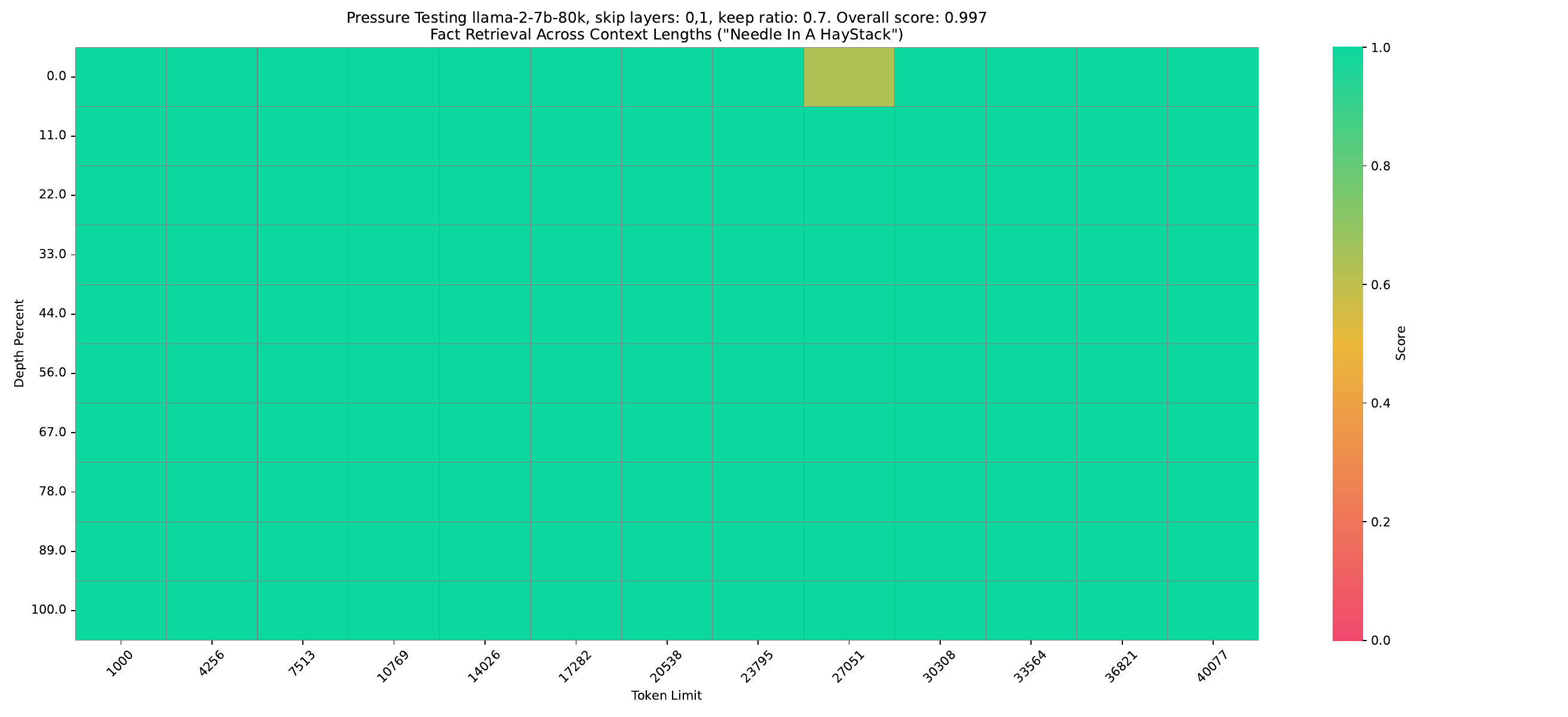}
    \captionsetup{width=0.9\linewidth}
    \caption{Llama-2-7b-80k, skip layer-0 and layer-1, compression ratio 30\%}
    % \label{fig:norm-attn-diff-llama-2-7b-longlora-32k-ft}
    \end{subfigure}

    \begin{subfigure}[t]{0.95\textwidth}
    \includegraphics[width=1\linewidth]{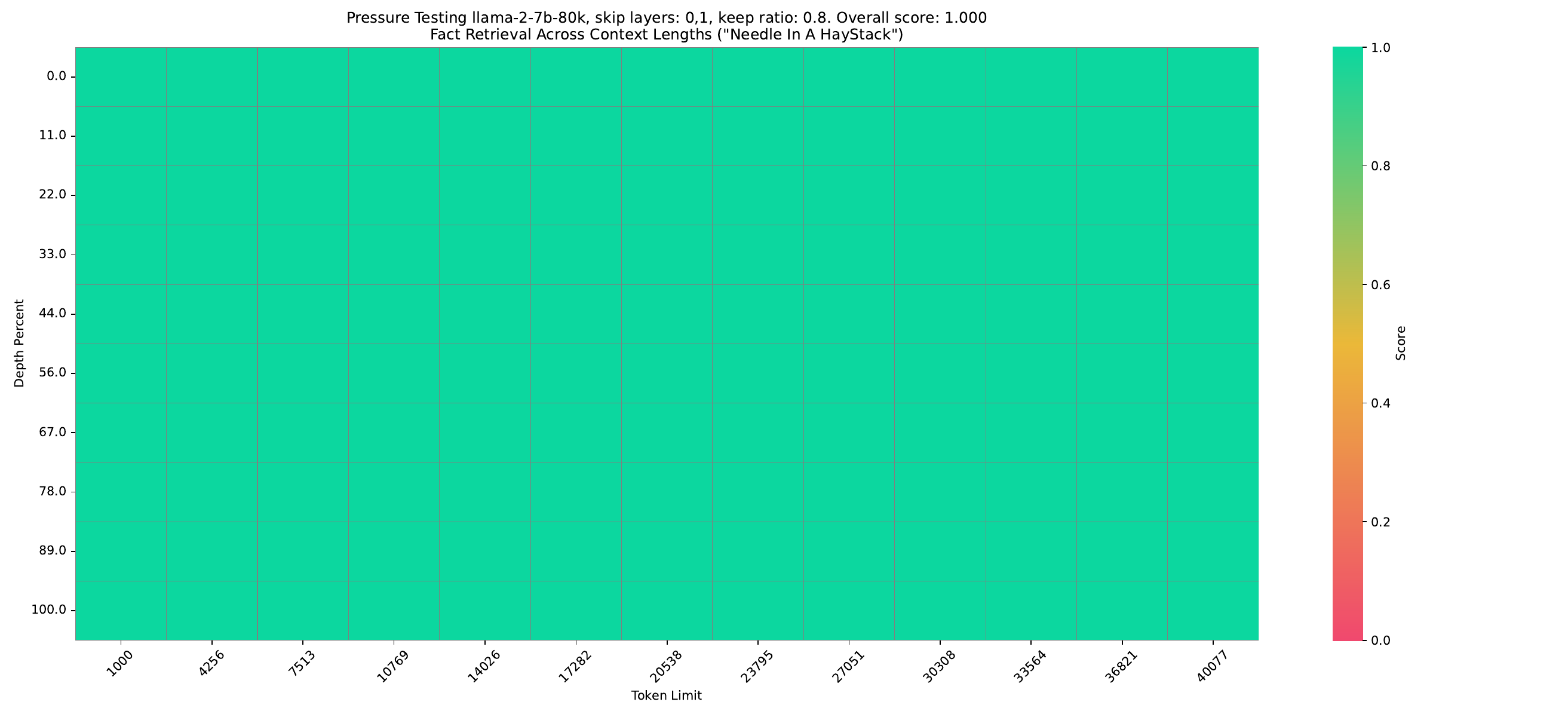}
    \captionsetup{width=0.9\linewidth}
    \caption{Llama-2-7b-80k, skip layer-0 and layer-1, compression ratio 20\%}
    % \label{fig:norm-attn-diff-llama-2-7b-longlora-32k-ft}
    \end{subfigure}
    \caption{Detailed results of Llama-2-7b-80k on the needle-in-a-haystack task.}
    \label{fig:llama-2-7b-80k-needle-detail}
\end{figure*}

\begin{figure*}
    \centering
    \begin{subfigure}[t]{0.95\textwidth}
    \includegraphics[width=1\linewidth]{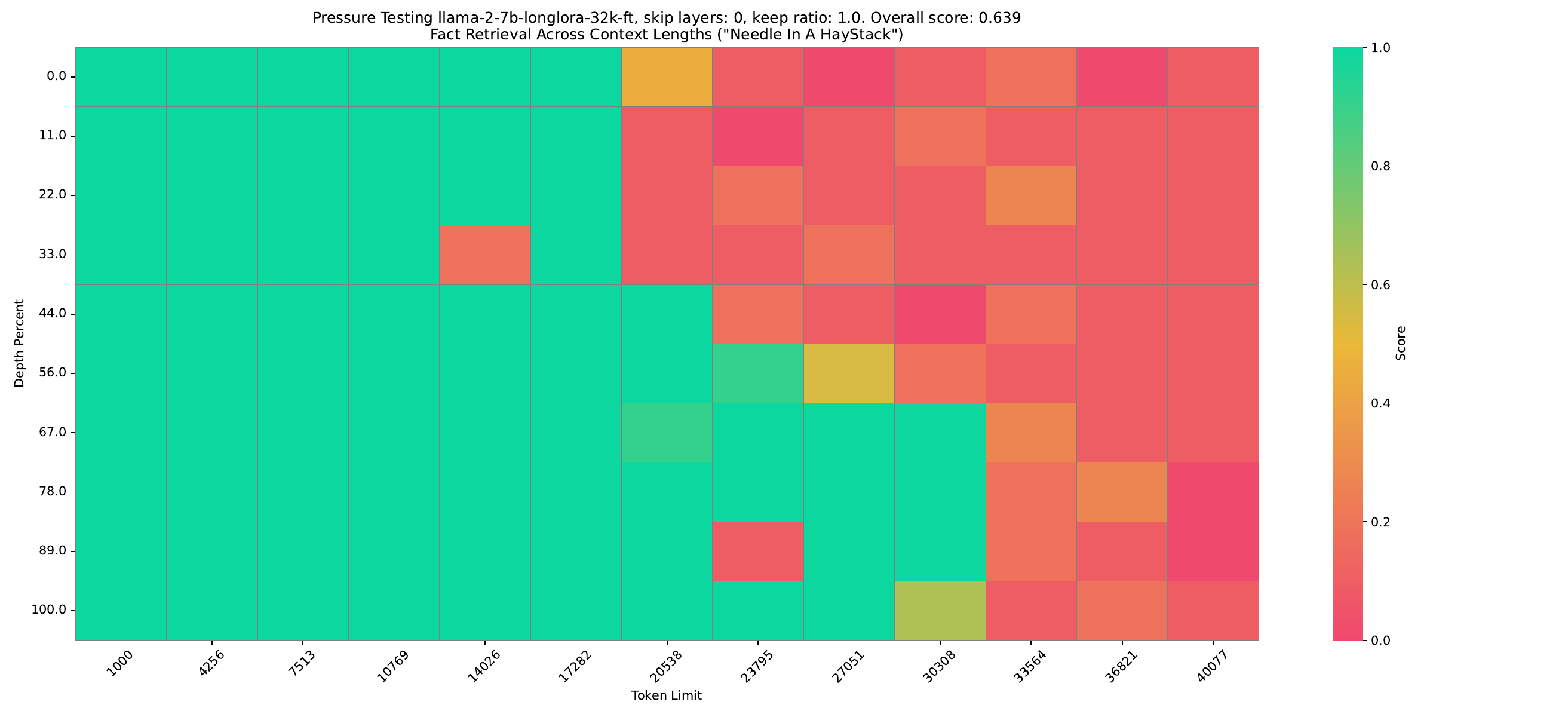}
    \captionsetup{width=0.9\linewidth}
    \caption{Llama-2-7b-longlora-32k-ft, without compression}
    % \label{fig:norm-attn-diff-llama-2-7b}
    \end{subfigure}

    \begin{subfigure}[t]{0.95\textwidth}
    \includegraphics[width=1\linewidth]{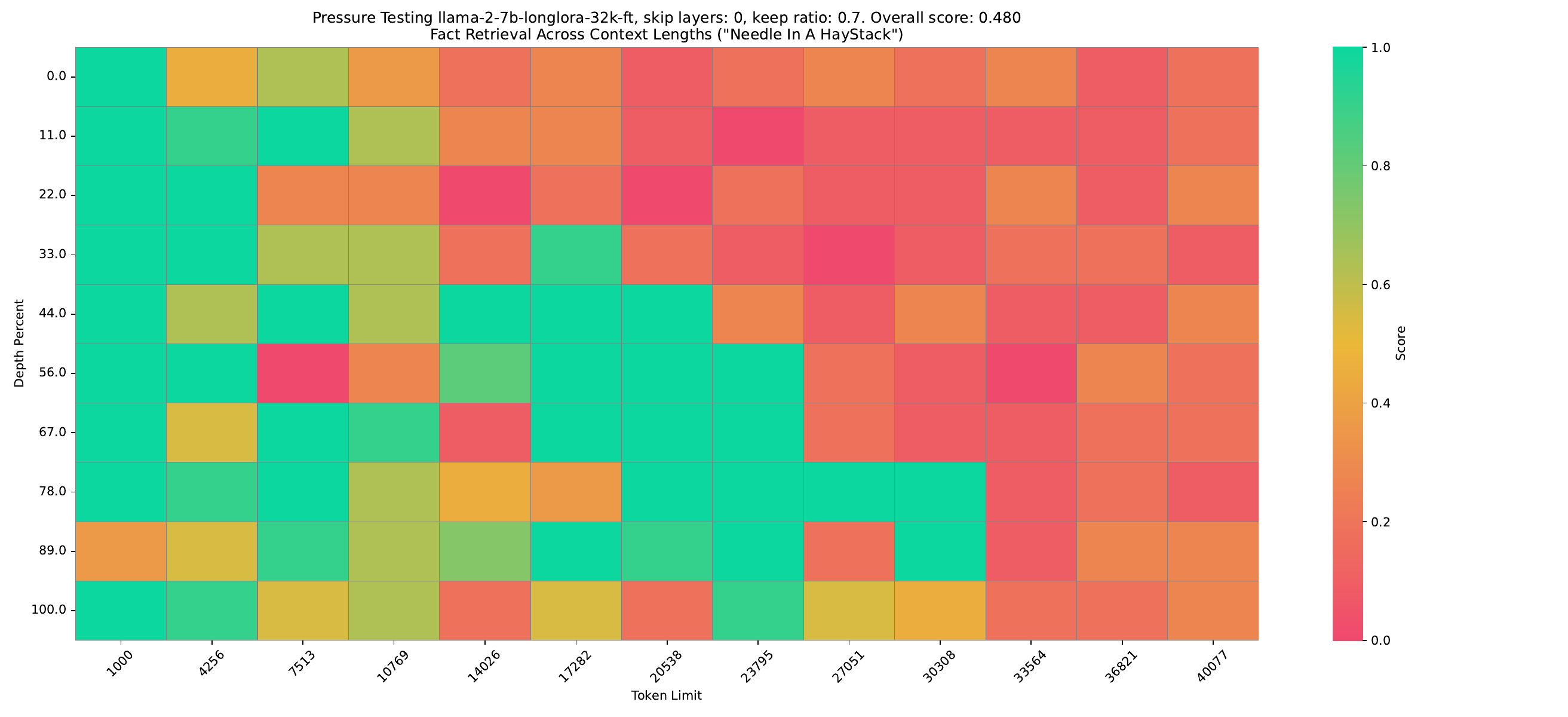}
    \captionsetup{width=0.9\linewidth}
    \caption{Llama-2-7b-longlora-32k-ft, skip layer-0, compression ratio 30\%}
    % \label{fig:norm-attn-diff-llama-2-7b-longlora-32k-ft}
    \end{subfigure}

    \begin{subfigure}[t]{0.95\textwidth}
    \includegraphics[width=1\linewidth]{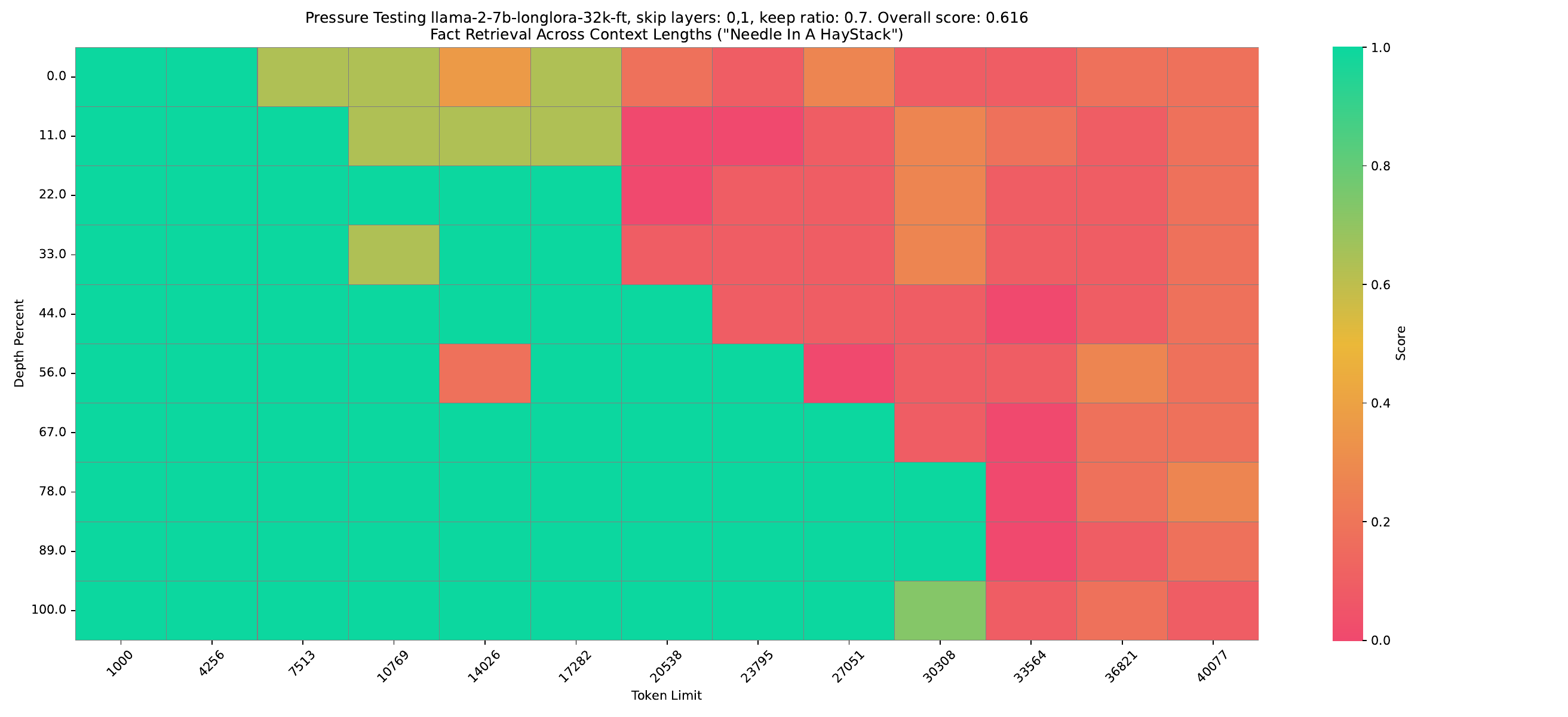}
    \captionsetup{width=0.9\linewidth}
    \caption{Llama-2-7b-longlora-32k-ft, skip layer-0 and layer-1, compression ratio 30\%}
    % \label{fig:norm-attn-diff-llama-2-7b-longlora-32k-ft}
    \end{subfigure}
    \caption{Detailed results of Llama-2-7b-longlora-32k-ft on the needle-in-a-haystack task.}
    \label{fig:llama-2-7b-longlora-32k-ft-needle-detail}
\end{figure*}

\begin{figure*}[!ht]
    \centering
    \begin{subfigure}[t]{0.45\textwidth}
    \includegraphics[width=1\linewidth]{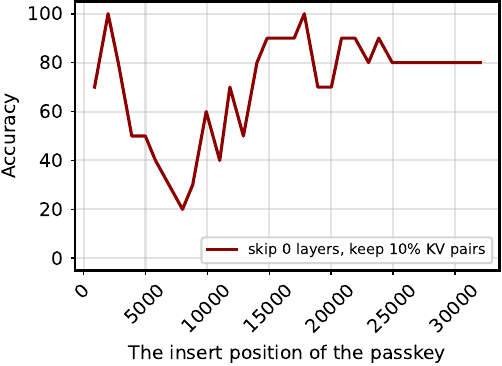}
    \captionsetup{width=0.9\linewidth}
    \caption{Llama-2-7b-80k, skip layer-0, compression ratio 90\%}
    % \label{fig:norm-attn-diff-llama-2-7b-longlora-32k-ft}
    \end{subfigure}
    \begin{subfigure}[t]{0.45\textwidth}
    \includegraphics[width=1\linewidth]{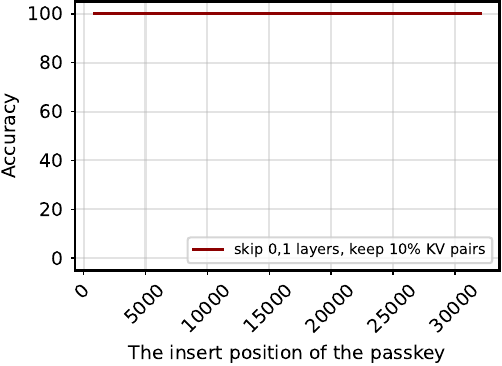}
    \captionsetup{width=0.9\linewidth}
    \caption{Llama-2-7b-80k, skip layer-0 and layer-1, compression ratio 90\%}
    % \label{fig:norm-attn-diff-llama-2-7b-longlora-32k-ft}
    \end{subfigure}
    \caption{Accuracy on the passkey retrieval. The $x$-axis presents the position of the passkey, and the $y$-axis presents the accuracy.}
    \label{fig:passkey-retrieval-detail}
\end{figure*}

\begin{figure*}[!ht]
    \centering
    \begin{subfigure}[t]{0.45\textwidth}
    \includegraphics[width=1\linewidth]{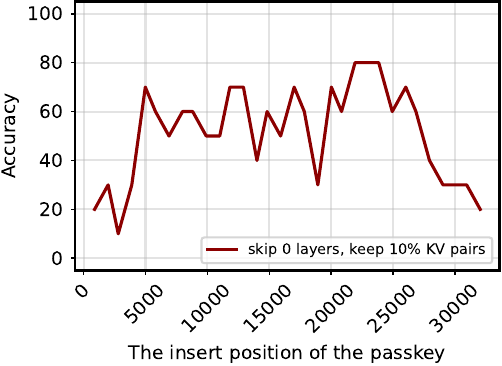}
    \captionsetup{width=0.9\linewidth}
    \caption{Llama-2-7b-longlora-32k-ft, skip layer-0, compression ratio 90\%}
    % \label{fig:norm-attn-diff-llama-2-7b-longlora-32k-ft}
    \end{subfigure}
    \begin{subfigure}[t]{0.45\textwidth}
    \includegraphics[width=1\linewidth]{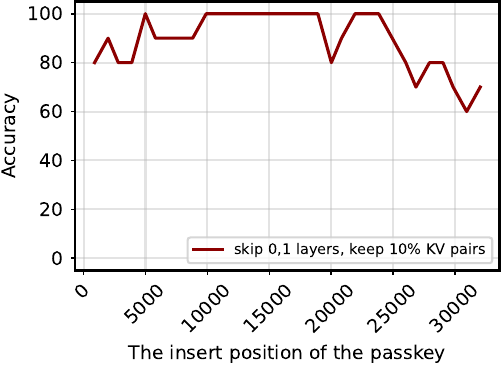}
    \captionsetup{width=0.9\linewidth}
    \caption{Llama-2-7b-longlora-32k-ft, skip layer-0 and layer-1, compression ratio 90\%}
    % \label{fig:norm-attn-diff-llama-2-7b-longlora-32k-ft}
    \end{subfigure}
    \caption{Accuracy on the passkey retrieval. The $x$-axis presents the position of the passkey, and the $y$-axis presents the accuracy.}
    \label{fig:passkey-retrieval-detail-longlora}
\end{figure*}

\clearpage
\subsection{Longbench Evaluation}
\begin{figure}[t]
    \centering
    \subfloat[]{\includegraphics[width=0.45\textwidth]{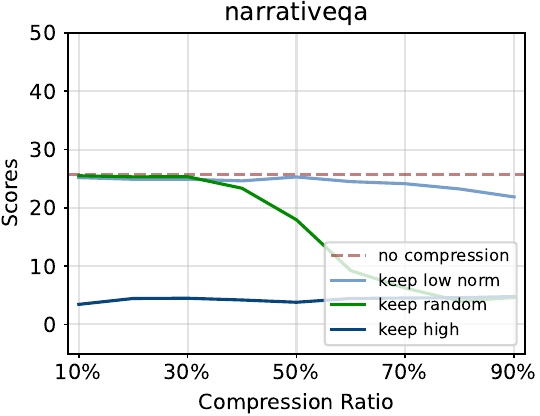}}\hfill
    \subfloat[]{\includegraphics[width=0.45\textwidth]{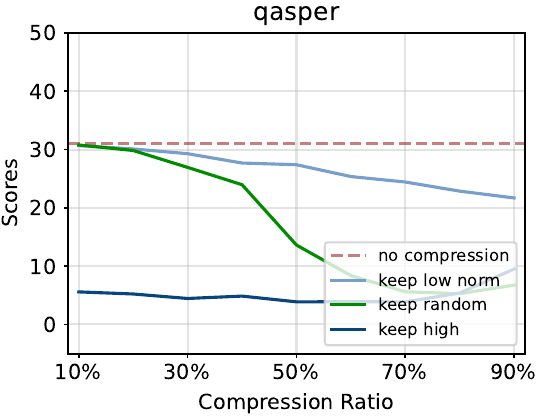}}\\
    \subfloat[]{\includegraphics[width=0.45\textwidth]{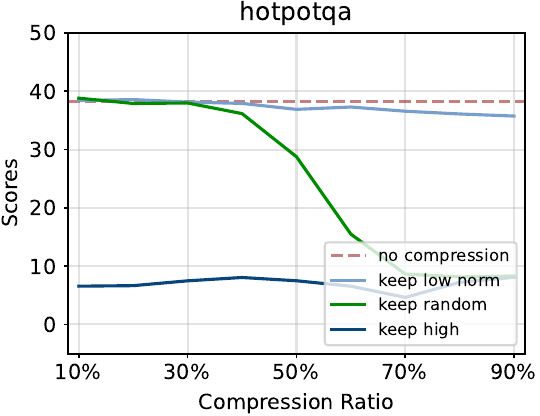}}\hfill
    \subfloat[]{\includegraphics[width=0.45\textwidth]{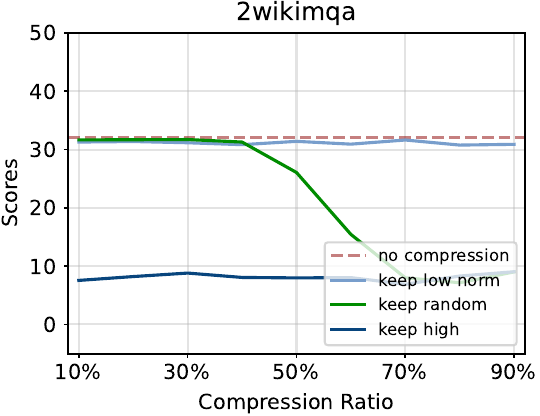}} \\
    \subfloat[]{\includegraphics[width=0.45\textwidth]{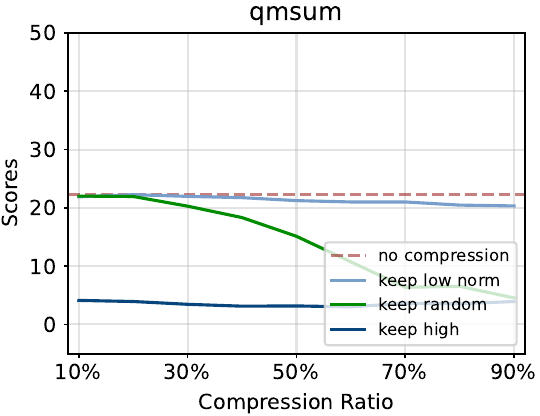}}\hfill
    % \subfloat[]{\includegraphics[width=0.45\textwidth]{figures/longbench/llama-2-7b-80k-longbench-gov_report.pdf}} \\
    % \subfloat[]{\includegraphics[width=0.45\textwidth]{figures/longbench/llama-2-7b-80k-longbench-average.pdf}}
    \subfloat[]{\includegraphics[width=0.45\textwidth]{figures/longbench/llama-2-7b-80k-longbench-average.pdf}}
    \caption{Evaluation results of Llama-2-7b-80k on long context tasks from Longbench, including narrativeqa and qasper, hotpotqa, 2wikimqa, and qmsum.}
    \label{fig:longbech_llama2}
\end{figure}

\begin{figure}[t]
    \centering
    \subfloat[]{\includegraphics[width=0.45\textwidth]{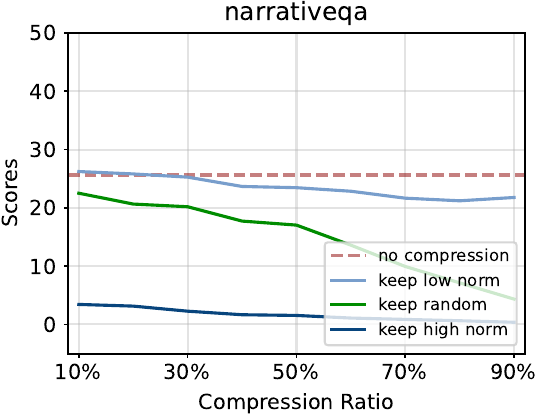}}\hfill
    \subfloat[]{\includegraphics[width=0.45\textwidth]{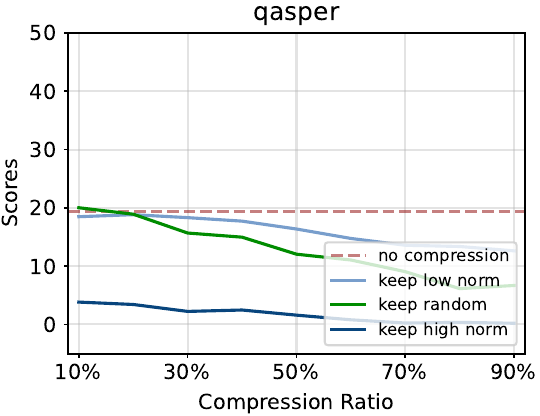}}\\
    \subfloat[]{\includegraphics[width=0.45\textwidth]{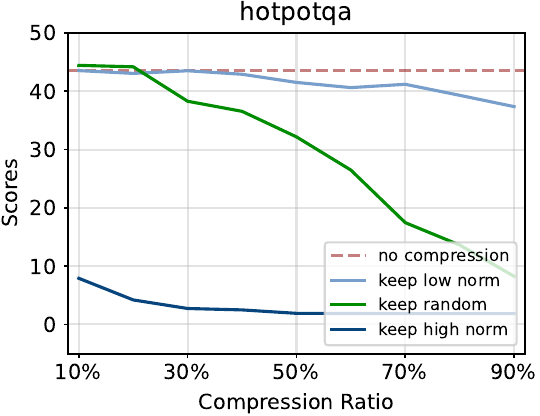}}\hfill
    \subfloat[]{\includegraphics[width=0.45\textwidth]{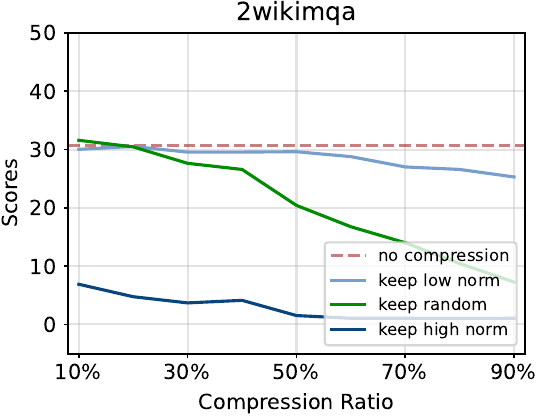}} \\
    \subfloat[]{\includegraphics[width=0.45\textwidth]{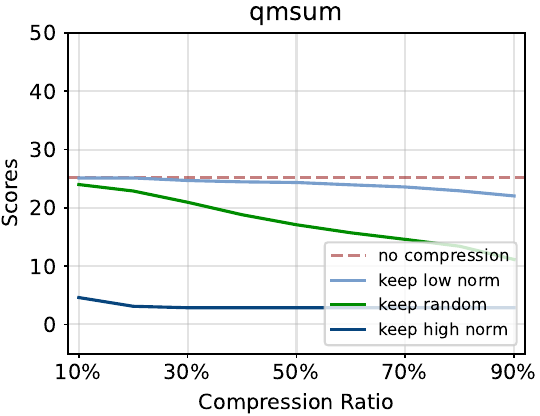}}\hfill
    \subfloat[]{\includegraphics[width=0.45\textwidth]{figures/longbench/Meta-Llama-3.1-8B-longbench-average.pdf}}
    \caption{Evaluation results of Llama-3.1-8B on long context tasks from Longbench, including narrativeqa and qasper, hotpotqa, 2wikimqa, and qmsum.}
    \label{fig:longbench_llama3}
\end{figure}

In this section we show detailed results from the LongBench dataset \citep{longbench-zhang-etal-2024}. In \Cref{fig:longbech_llama2} we show results for Llama2-80k, while in \Cref{fig:longbench_llama3} we show results for the long context model Llama3.1-8b.
 
\section{More Visualizations}
\label{sec:more_visualizations}

\newpage
\thispagestyle{empty} % Removes the page number
\newgeometry{left=0cm, right=0cm, top=0cm, bottom=0cm} % Set margins to 0 for full page image

\begin{figure*}
    \centering
    \includegraphics[width=0.85\textwidth]{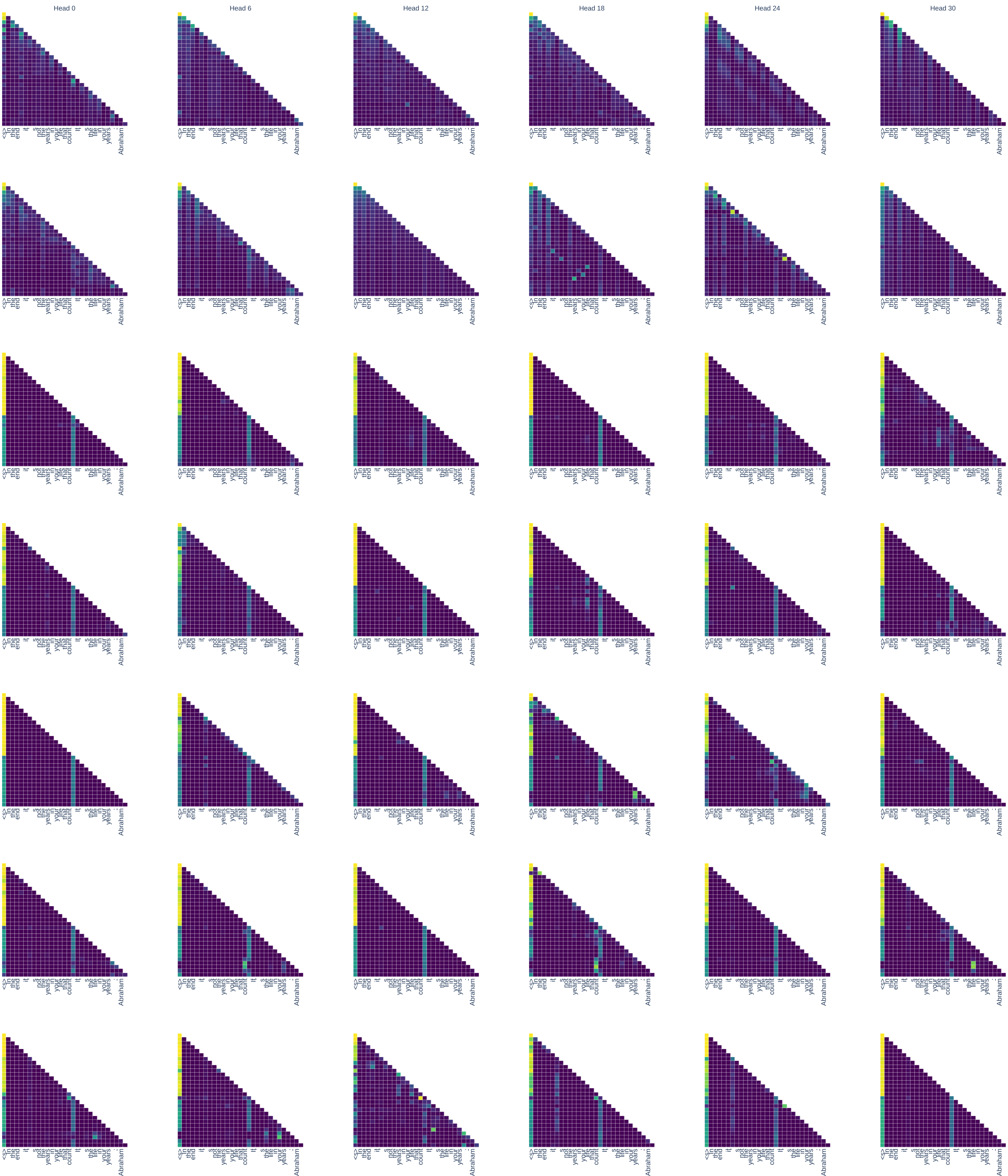}
    \caption{Attention maps in Llama2-7B}
    \label{fig:chunk_4_attns}
\end{figure*}
\begin{figure*}
    \centering
    \includegraphics[width=0.85\textwidth]{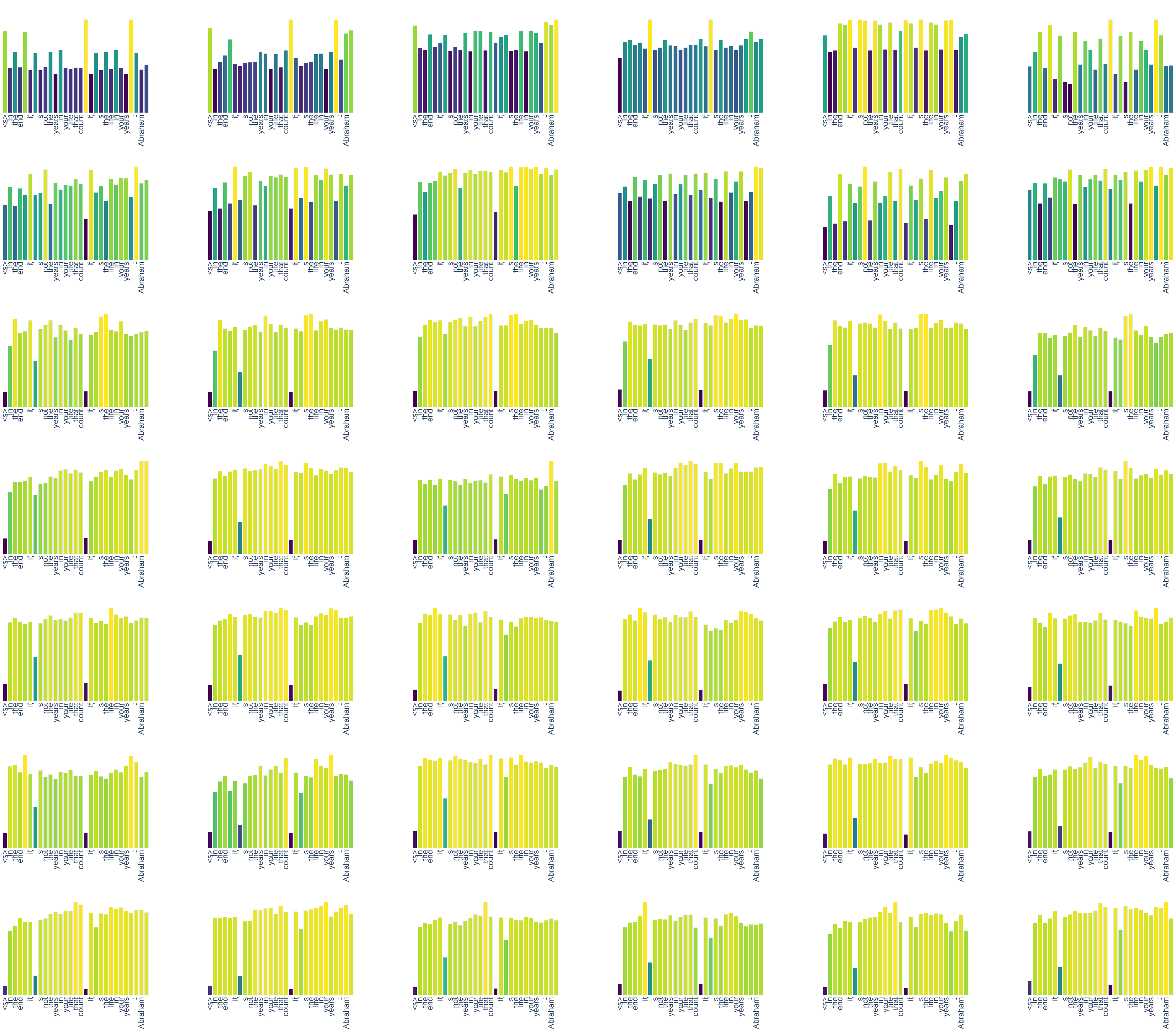}
    \caption{Norms of KV cache tokens in Llama2-7B}
    \label{fig:chunk_4_norms}
\end{figure*}
\begin{figure*}
    \centering
    \includegraphics[width=0.85\textwidth]{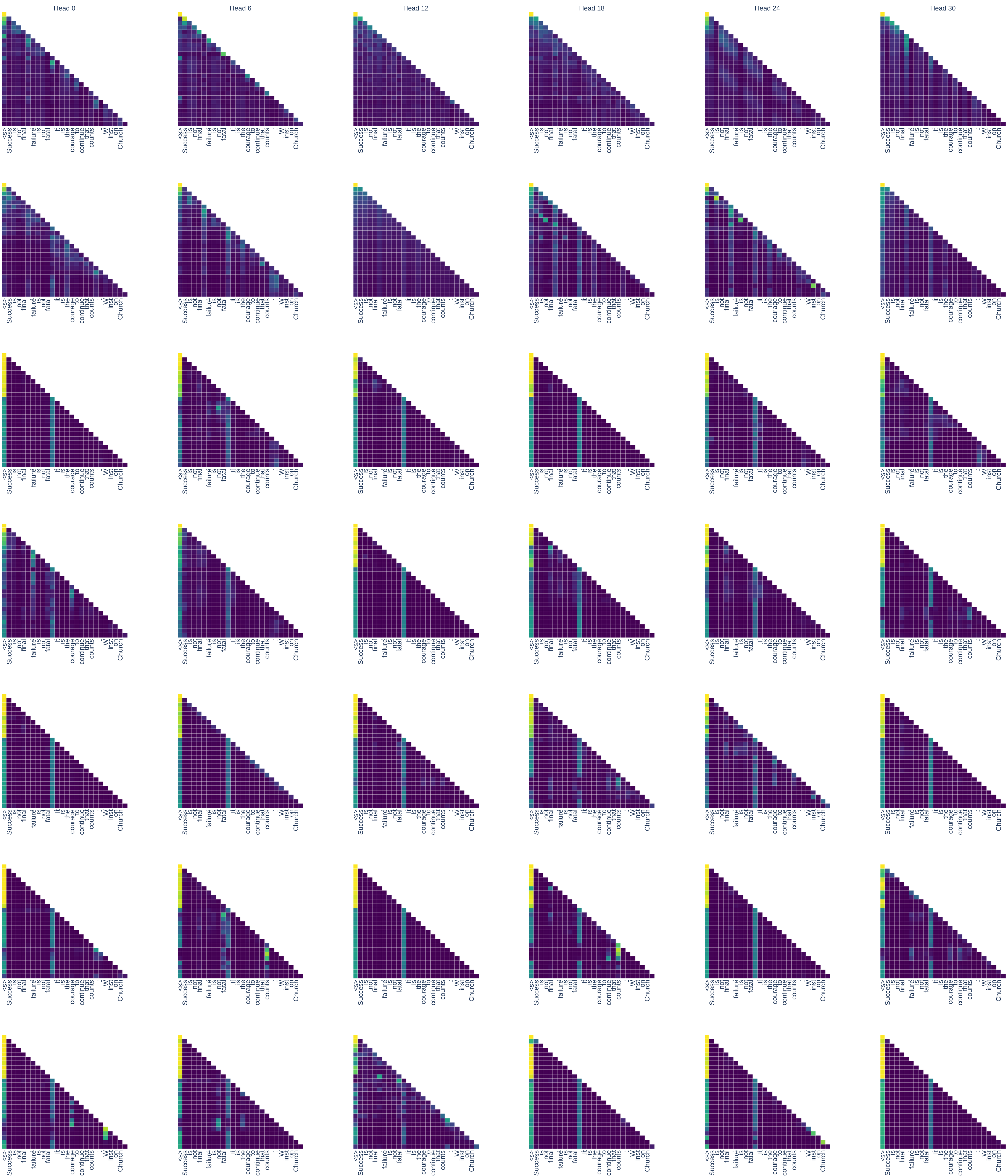}
    \caption{Attention maps in Llama2-7B}
    \label{fig:chunk_5_attns}
\end{figure*}

\begin{figure}
    \centerline{
    \includegraphics[width=0.85\textwidth]{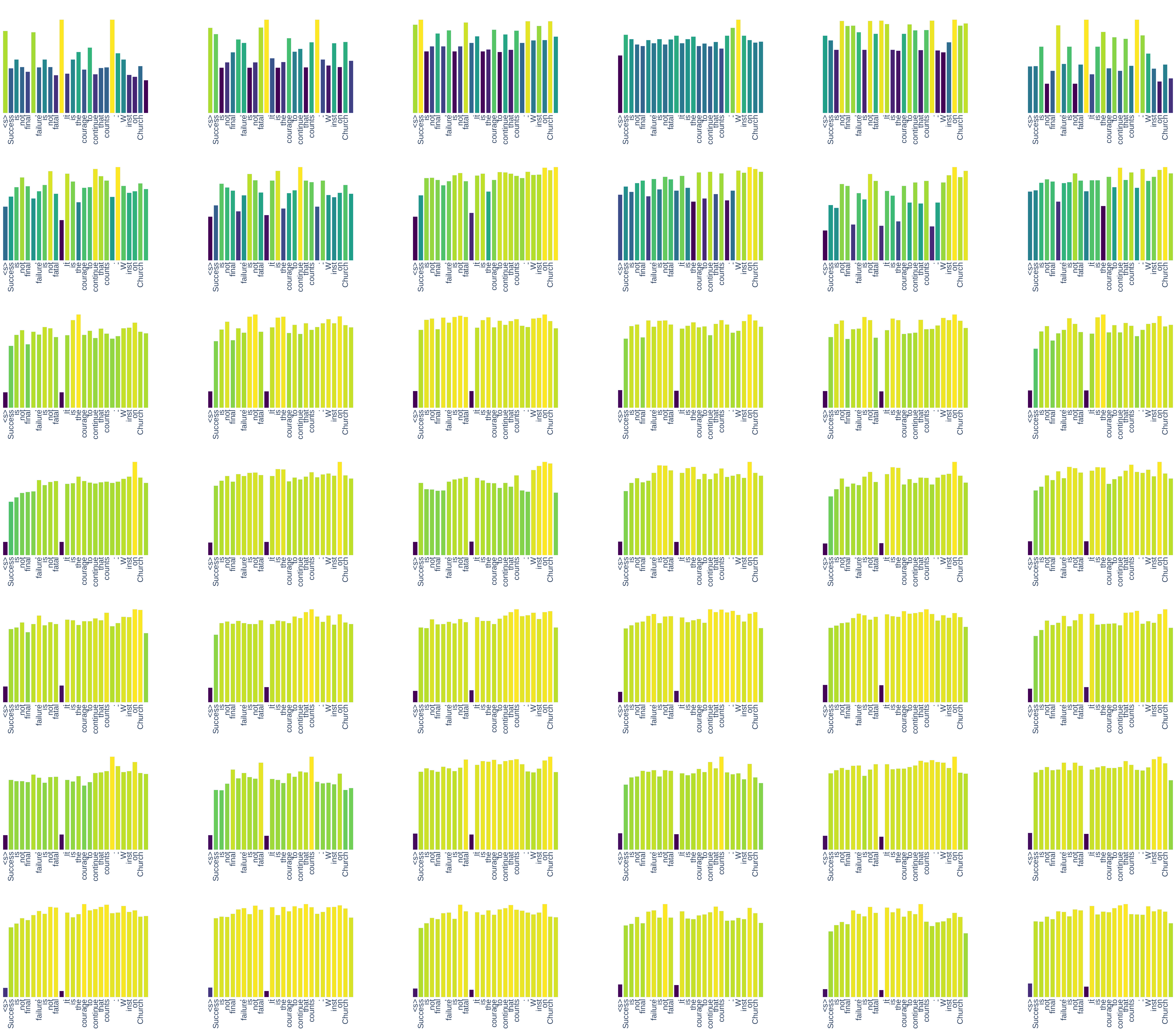}}
    \caption{Norms of KV cache tokens in Llama2-7B}
    \label{fig:chunk_5_norms}
\end{figure}
\begin{figure*}
    \centering
    \includegraphics[width=0.65\textwidth]{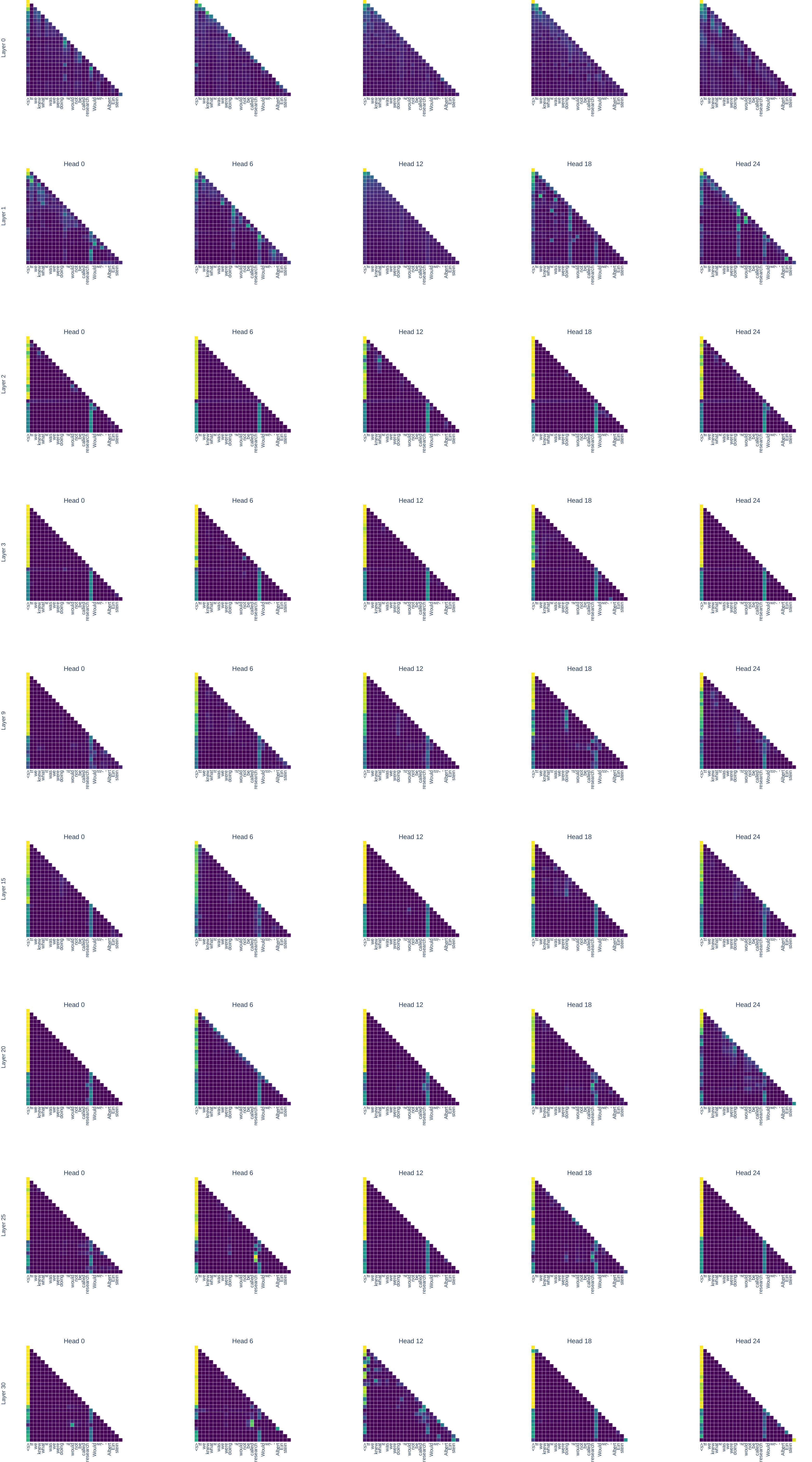}
    \caption{Attention maps in Llama2-7B}
    \label{fig:chunk_3_attns}
\end{figure*}
\begin{figure*}
    \centering
    \includegraphics[width=0.65\textwidth]{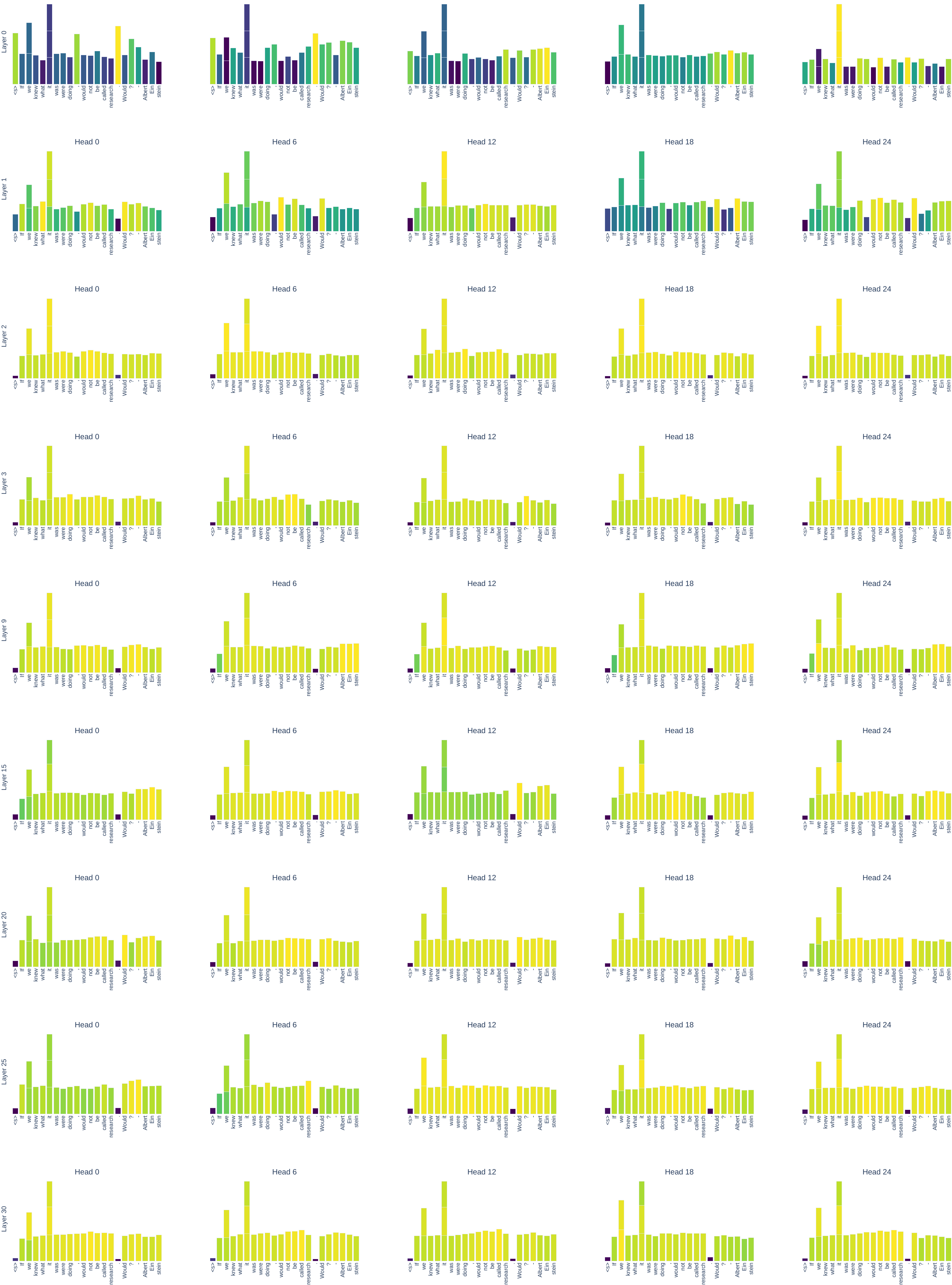}
    \caption{Norms of KV cache tokens in Llama2-7B}
    \label{fig:chunk_3_norms}
\end{figure*}

\restoregeometry % Restore the original margins

\section{Additional token embeddings plots}
\label{sec:more_token_emb}
\begin{figure}[b]
    \centering
    \subfloat[]{\includegraphics[width=0.4\textwidth]{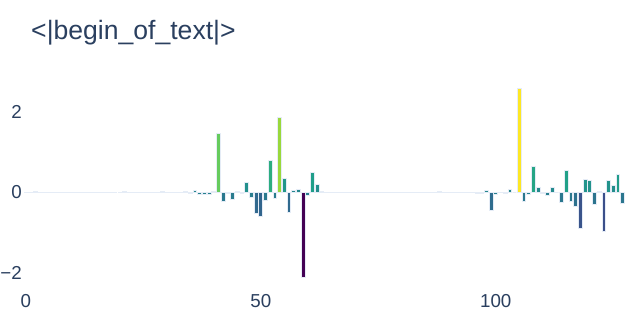}} \hfill
    \subfloat[]{\includegraphics[width=0.4\textwidth]{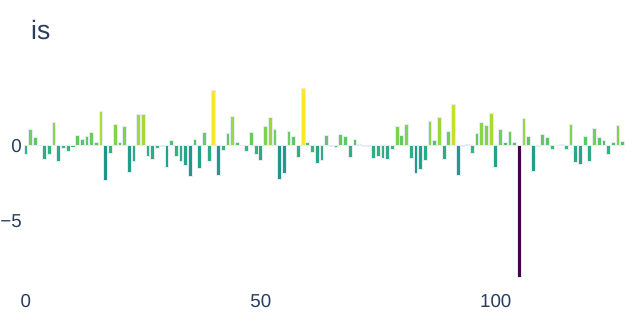}} \\
    \subfloat[]{\includegraphics[width=0.4\textwidth]{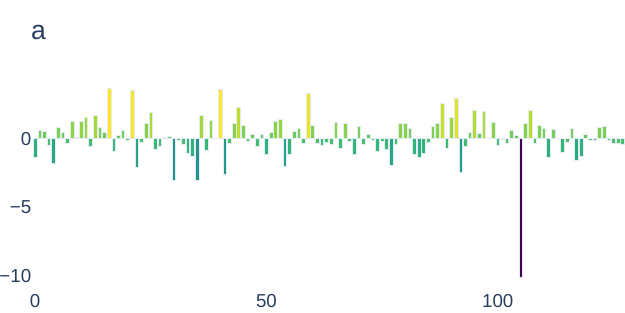}} \hfill
    \subfloat[]{\includegraphics[width=0.4\textwidth]{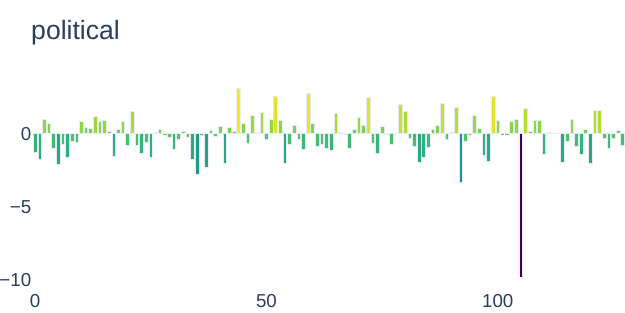}} \\
    \subfloat[]{\includegraphics[width=0.4\textwidth]{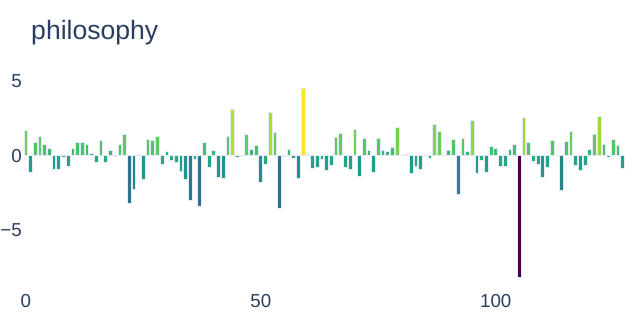}} \hfill
    \subfloat[]{\includegraphics[width=0.4\textwidth]{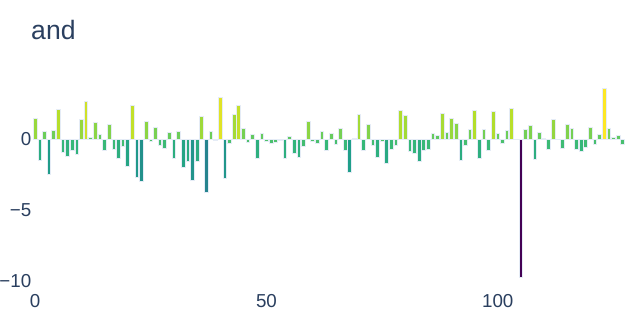}} \\
    \caption{Key projections of Llama3-8b of the bos $|begin of text|$ token vs other tokens. Each value represents the activation in a specific dimension for the embedding of the key projection. We found similar patterns across almost all heads and layers and in multiple texts.}
    \label{fig:llam3_token_emb}
\end{figure}

We show in \Cref{fig:llam3_token_emb} some additional figure that represent Llama3-8b token embeddings sparsity. 
\section{Experimental setup}
\label{sec:exp_setup}
In all experiments, we used the HuggingFace library and did not change the model's default hyperparameters. 
For language modelling, results are averaged across 50 samples.
The \cref{fig:norm-attn-diff-layer-head} and \cref{fig:norm-attn-diff} are the average results of $1024$ examples with a chunk size of $1024$ using Wikipedia.
%
%
% no need for checklist 
%\clearpage
%\input{appendix/checklist}

\end{document}